\documentclass[sigconf]{acmart}
\AtBeginDocument{%
  }

\copyrightyear{2026}
\acmYear{2026}
\setcopyright{cc}
\setcctype{by}
\acmConference[MM '26]{Proceedings of the 34th ACM International Conference on Multimedia}{November 10--14, 2026}{Rio de Janeiro, Brazil.}
\acmBooktitle{Proceedings of the 34th ACM International Conference on Multimedia (MM '26), November 10--14, 2026, Rio de Janeiro, Brazil}
\acmISBN{979-8-4007-2213-4/2026/11}
\acmDOI{10.1145/3767308.3835230}

\usepackage{amsfonts}
\usepackage{multirow}
\usepackage{subfigure}
\usepackage{colortbl}

\usepackage{pifont}

\usepackage[most]{tcolorbox}
\usepackage{amsmath}

\newcommand{\cmark}{\textcolor{green!70!black}{\ding{51}}}
\newcommand{\xmark}{\textcolor{red}{\ding{55}}}  

\begin{document}

\title{Peak-End-Net: A Peak-End Rule Inspired Framework for Generalizable Video Aesthetic Assessment}

\author{Geng Li}
\authornote{Both authors contributed equally to this work.}
\email{xiaofeng.lg@alibaba-inc.com}
\affiliation{
  \institution{AMAP, Alibaba Group}
  \city{Beijing}
  \country{China}
}

\author{Haiwen Li}
\authornotemark[1]
\authornote{Work done during the internship at AMAP, Alibaba Group.}
\email{lihaiwen.lhw@alibaba-inc.com}
\affiliation{
  \institution{AMAP, Alibaba Group}
  \city{Beijing}
  \country{China}
}

\author{Rui Chen}
\email{chenrui.chen@alibaba-inc.com}
\affiliation{
  \institution{AMAP, Alibaba Group}
  \city{Beijing}
  \country{China}
}

\author{Jing Tang}
\email{guangyu.tj@alibaba-inc.com}
\affiliation{
  \institution{AMAP, Alibaba Group}
  \city{Beijing}
  \country{China}
}

\author{Lei Sun}
\email{ally.sl@alibaba-inc.com}
\affiliation{
  \institution{AMAP, Alibaba Group}
  \city{Beijing}
  \country{China}
}

\author{Xiangxiang Chu}
\email{chuxiangxiang.cxx@alibaba-inc.com}
\affiliation{
  \institution{AMAP, Alibaba Group}
  \city{Beijing}
  \country{China}
}

\renewcommand{\shortauthors}{Geng Li et al.}

\begin{abstract}

Video aesthetic assessment (VAA) aims to predict how aesthetically pleasing a video is, yet remains far less explored than other visual assessment tasks. Its progress is hindered not only by the scarcity of large-scale benchmarks, but also by the intrinsic subjectivity of aesthetic judgment, which is shaped by human perception. In this paper, we revisit VAA from a psychological perspective and propose \textit{Peak-End-Net}, a lightweight and interpretable framework inspired by the \textit{peak-end rule}, which suggests that people tend to judge a temporal experience mainly according to its salient moments and the ending. Building on this intuition, we first transfer knowledge from image aesthetic assessment (IAA) to VAA by introducing a pretrained IAA head to produce frame-wise aesthetic priors, which serve as surrogate signals for identifying aesthetically salient moments and guiding \textit{peak-end rule}-based temporal aggregation. To further capture how a video evolves aesthetically over time, we design an aesthetic rhythm encoder that models temporal progression beyond isolated moments. Additionally, we refine the overall assessment through a dynamic gated fusion mechanism to improve robustness under distribution shift. Our method is built on a frozen vision transformer (ViT) and requires only a small number of trainable parameters, making it scalable and parameter-efficient. Extensive experiments on two existing VAA benchmarks, including in-domain evaluation on VADB and cross-domain testing on DIVIDE-3K, demonstrate that our approach achieves state-of-the-art performance, affirming the value of psychologically grounded modeling for VAA. Our code and models are available at \url{https://github.com/AMAP-ML/Peak-End-Net}.

\end{abstract}



\maketitle

\section{Introduction}
The explosive growth of online video platforms and AI-generated media has made video a ubiquitous form of visual communication. Beyond semantic content and technical fidelity, viewers increasingly care about whether a video is aesthetically pleasing, that is, whether its composition, lighting, color, motion, and overall visual flow evoke a favorable impression. This demand has rendered video aesthetic assessment (VAA)~\cite{vadb,divide} an important yet still underexplored problem in multimedia research.

\begin{figure}[t]
\centering
\includegraphics[width=\columnwidth]{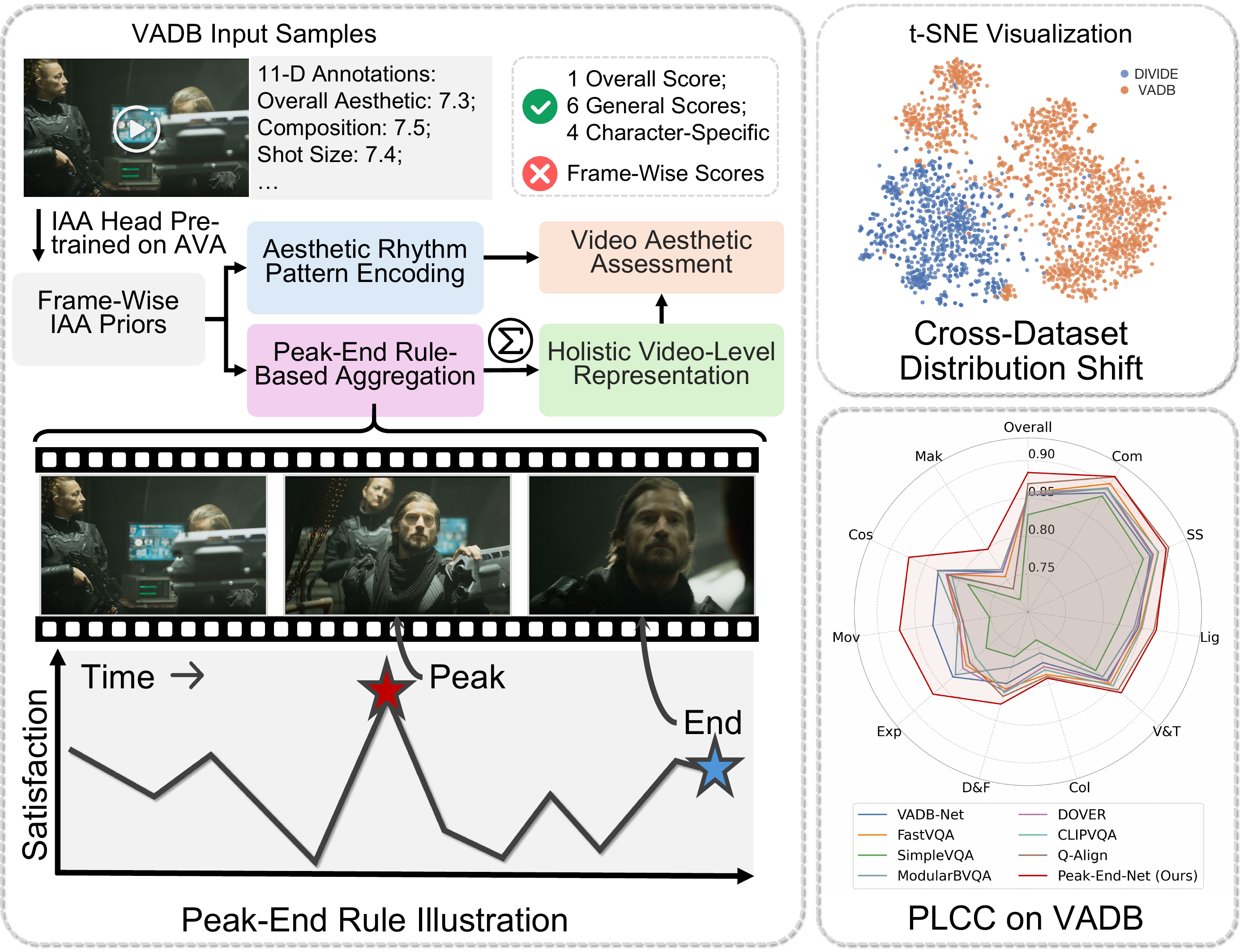}
\caption{Illustration of our motivation. VAA is inherently subjective and susceptible to cross-dataset distribution shift. We draw inspiration from the psychologically grounded \textit{peak-end rule}~\cite{peak-end}, leverage frame-wise IAA priors to identify aesthetically salient moments and encode aesthetic rhythm patterns to capture temporal evolution for generalizability.}
\Description{Motivation for video aesthetic assessment: aesthetic judgment is subjective and shifts across datasets, motivating peak-end rule based modeling with frame-wise aesthetic priors and rhythm patterns.}
\label{fig:intro}
\end{figure}

Despite its practical relevance, progress in VAA remains hindered by two long-standing bottlenecks. The first is data scarcity. Compared with benchmarks for mature video understanding tasks~\cite{ucf101,hmdb,mvbench,videounderstand}, VAA benchmarks are still limited in both scale and annotation richness.  To date, VADB~\cite{vadb} is the only large-scale benchmark specifically dedicated to VAA, offering 10,490 videos annotated by professionals with overall scores, attribute scores, comments, and tags. The second bottleneck is methodological limitations. Few VAA methods~\cite{professionalvideo,telefonica,vadb,divide} have been proposed, and many of them largely inherit designs from related video assessment tasks rather than being tailored to the distinctive nature of aesthetic judgment.

A key reason why VAA is intrinsically challenging is that it is highly subjective. Unlike scores in more objective visual assessment problems, aesthetic scores are ultimately assigned by humans and are therefore shaped not only by visual content itself, but also by how people summarize and remember their viewing experience. However, current methods typically formulate VAA as a generic video-level regression problem, without explicitly accounting for the cognitive mechanisms that may underlie human aesthetic judgment. As a result, they often fall short of interpretability and may generalize poorly under distribution shift.

Motivated by this observation, we revisit VAA through the lens of the \textit{peak-end rule}~\cite{peak-end}, a classic theory in psychology stating that people tend to judge an experience primarily by its most intense moments and its ending, rather than by uniformly averaging every moment. This principle is especially plausible for video aesthetics, where a few striking moments and the final impression may disproportionately shape the overall score. Accordingly, we propose to explicitly harness the \textit{peak-end rule} for the temporal aggregation of frame-level visual features, so as to form a holistic video-level representation that better aligns with human aesthetic judgment, as illustrated in Fig.~\ref{fig:intro}. Instead of treating all frames equally, our model emphasizes aesthetically salient moments and the ending segment during aggregation, yielding a more psychologically grounded and interpretable representation for VAA. To make this idea feasible, we further introduce frame-wise aesthetic priors, with image aesthetic assessment (IAA)~\cite{iaa,iaaphoto} serving as a natural precursor. While VAA lacks dense frame-level annotations, IAA has benefited from large-scale datasets such as AVA~\cite{ava} and has developed transferable aesthetic knowledge. Building on this, we employ an IAA head pretrained on AVA to infer frame-wise aesthetic scores, which serve as surrogate signals for identifying salient moments and guiding \textit{peak-end}-based aggregation. In this way, image-level aesthetic knowledge is transferred to video-level assessment without requiring manual frame-level annotation.

Beyond isolated peaks and ends, we argue that video aesthetics also depends on how the aesthetic experience evolves over time. Two videos may share similar salient moments yet evoke different impressions because their aesthetic progression follows divergent trajectories. To this end, we introduce aesthetic rhythm pattern encoding, which models the temporal variation structure of frame-wise aesthetic cues, including trends, fluctuations, and overall progression. 

Moreover, to improve generalization, especially under cross-dataset distribution shift, we refine the predicted overall score via a dynamic gated fusion mechanism. Specifically, it adaptively integrates the holistic judgment from the VAA branch with the transferable insight from the IAA branch. This refinement is advantageous because the VAA prediction may inherit dataset-specific bias, whereas the IAA prior is learned from large-scale image aesthetics data and thus offers stronger transferability. 

Our framework is lightweight by design. It builds on a frozen ViT backbone~\cite{vit,clip,explain} and incorporates only a small number of learnable modules. At the same time, it is psychologically grounded: rather than fabricating an opaque temporal aggregation strategy, it adopts an explicit cognitive prior that affords a compelling degree of interpretability. In summary, our contributions are fourfold:
\begin{itemize}
    \item We present a psychologically grounded framework for VAA, which explicitly incorporates the \textit{peak-end rule} into temporal aggregation for interpretable video aesthetic modeling.
    \item We bridge IAA and VAA through frame-wise aesthetic priors, using a pretrained IAA head to guide \textit{peak-end} aggregation without requiring frame-level video annotations.
    \item We introduce aesthetic rhythm pattern encoding and dynamic gated fusion, which capture temporal aesthetic progression and refine overall assessment for better robustness.
    \item Extensive experiments on two existing VAA benchmarks demonstrate the effectiveness of our method in both in-domain evaluation and cross-dataset generalization.
\end{itemize}

\section{Related Work}
\textbf{Video Aesthetic Assessment (VAA).} Image aesthetic assessment (IAA)~\cite{iaa,iaaphoto,ava,lapis} provides the foundation for VAA, but directly transferring image-based methods to videos is nontrivial because video aesthetics depends not only on frame-level appearance, such as composition, lighting, and color, but also on temporal cues like motion and rhythm. Early VAA studies~\cite{photovaa,telefonica,professionalvideo} mainly relied on handcrafted spatial and motion features, while later works~\cite{divide,avaq6000} gradually shifted to deep models, often borrowing architectures from video quality assessment (VQA)~\cite{vqa,fastvqa,simplevqa,modularbvqa}. Progress in this area has been constrained by the scarcity of large-scale and richly annotated datasets. Early datasets, such as the Telefonica dataset~\cite{telefonica}, VAQ700~\cite{vqa700}, and AVAQ6000~\cite{avaq6000}, were relatively limited in scale or annotation granularity and were not specifically tailored to VAA, while DIVIDE-3K~\cite{divide} later provided more diverse ratings, including an aesthetic dimension. More recently, VADB~\cite{vadb} advanced the field by introducing the largest video aesthetic dataset to date, with professional and multi-dimensional annotations including overall scores, attribute scores, comments, and tags. Correspondingly, VADB-Net~\cite{vadb} showed that multimodal pretraining with language comments and technical tags can effectively improve VAA.

\textbf{Vision-Language Models (VLMs)} have become a powerful paradigm for transferable multimodal representation learning. Early VLMs are typically based on dual encoders, such as CLIP~\cite{clip}, ALIGN~\cite{align}, BLIP~\cite{blip}, and SigLIP~\cite{siglip,siglip2}, which learn aligned visual and textual embeddings through large-scale contrastive training~\cite{moco,simclr}. Building on these models, prior work explored prompt learning and adapter-based tuning, and extended image-level VLMs to video tasks through models such as X-CLIP~\cite{xclip}, CLIP4Clip~\cite{clip4clip}, ActionCLIP~\cite{actionclip}, and VideoCLIP~\cite{videoclip}. More recently, VLMs have also been introduced into perceptual assessment, where methods such as PTM-VQA~\cite{ptmvqa}, CLIPVQA~\cite{clipvqa}, and Q-CLIP~\cite{qclip} affirm the value of language-guided representations for quality-related prediction. Beyond dual-encoder contrastive models, the field has increasingly shifted toward multimodal large language models (MLLMs)~\cite{qwen3vl,internvl,gpt4}, which pair visual encoders with large language models to enable open-ended reasoning~\cite{qinsight, vqr1, chen2025finger, li2025next} and natural-language explanations for visual scoring~\cite{qalign, ling2025vmbench, feng2025narrlv, xie2026qhawkeye}.

\begin{figure*}
\centering
\includegraphics[width=\textwidth]{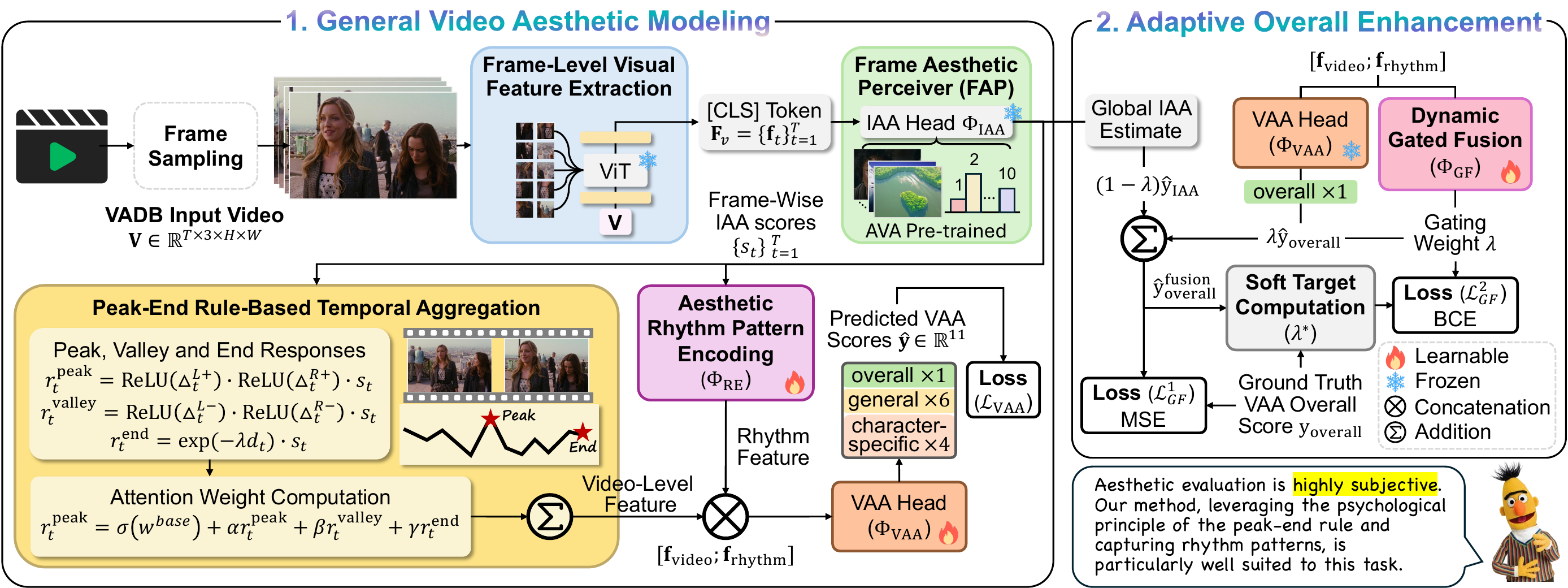}
\caption{Overview of our method. Given sampled video frames, the system transfers frame-wise aesthetic priors from an IAA head pretrained on AVA~\cite{ava}, performs \textit{peak-end rule}-based temporal aggregation and aesthetic rhythm encoding for psychologically grounded VAA modeling, and further refines the overall score via dynamic gated fusion for improved generalization.}
\Description{Main framework of the paper}
\label{fig:main}
\end{figure*}

\section{Methodology}
An overview of the entire framework is visualized in Fig.~\ref{fig:main}. We first introduce the preliminaries in Sec.~\ref{3.1}. Then, Secs.~\ref{3.2} and~\ref{3.3} detail the two stages, including their forward procedures and loss functions, respectively. Finally, Sec.~\ref{3.4} presents the inference process.

\subsection{Preliminary}
\label{3.1}
\textbf{Image Aesthetic Assessment.} IAA aims to predict the aesthetic quality of an image, typically as a score distribution or a mean score. In AVA~\cite{ava}, each image is annotated with votes over 10 discrete aesthetic levels, forming a ground-truth distribution over scores from 1 to 10. Accordingly, an IAA model outputs 10-dimensional logits, which are normalized into a probability distribution, and the final score is computed as the expectation over the 10 levels.

\textbf{Video Aesthetic Assessment.}
Given a video, VAA aims to predict its overall aesthetic quality and, when available, multiple attribute-specific scores. In VADB~\cite{vadb}, each video is annotated with up to 11 dimensions, including 1 overall score, 6 general scores, and 4 character-specific scores defined only for videos containing human subjects. Thus, non-character videos contain only 7 valid dimensions. All scores range from 0 to 10 and are normalized to \([0,1]\) during training. A VAA model predicts an 11-dimensional output vector followed by a sigmoid function.

$\star$ The key difference is that IAA provides frame-independent aesthetic assessment but lacks temporal modeling, whereas VAA targets holistic video aesthetics but usually has no frame-wise annotations. This motivates leveraging image-level aesthetic priors for video-level aesthetic modeling.

\subsection{General Video Aesthetic Modeling}
\label{3.2}

\textbf{Frame-Level Visual Feature Extraction.}
Given an input video \( \mathbf{V} \in \mathbb{R}^{T \times 3 \times H \times W} \), where \(T\) denotes the number of sampled frames, we employ a frozen vision transformer (ViT)~\cite{vit,clip} to encode each frame independently. Specifically, each frame is first partitioned into non-overlapping patches and linearly projected into patch embeddings. A learnable \([CLS]\) token is then prepended to the patch sequence, and the resulting token sequence of each frame is fed into the transformer encoder~\cite{transformer}. For each frame, we take the \([CLS]\) token and apply the visual projection of CLIP~\cite{clip} to obtain its \(d\)-dimensional visual representation, producing the frame-level feature sequence:
\begin{equation}
\mathbf{F}_v = \{\mathbf{f}_t\}_{t=1}^{T} \in \mathbb{R}^{T \times d}.
\end{equation}

$\star$ This design offers two advantages: \ding{182} the frozen ViT backbone provides strong visual priors for capturing fine-grained appearance details and aesthetics-related cues; \ding{183} it is parameter-efficient, while the frozen ViT is readily reusable, offering strong scalability for extension to subsequent modules and related tasks.

\textbf{Frame Aesthetic Perceiver.} Since a video is composed of a sequence of frames, the aesthetic quality of individual frames naturally contributes to the overall video aesthetic judgment. Motivated by this, we introduce the Frame Aesthetic Perceiver (FAP) that interfaces with \( \mathbf{F}_v \) and predicts an aesthetic score for each frame. We refer to this score as the IAA score, which serves as a frame-wise aesthetic prior for subsequent video-level modeling.

Concretely, we employ an IAA head \( \Phi_{\text{IAA}}(\cdot) \), a multi-layer perceptron (MLP), to map each frame feature to $K$-dimensional logits\footnote{$K$ is set to 10 to align with the 10 aesthetic levels in the AVA~\cite{ava} dataset.}, corresponding to discrete aesthetic levels from 1 to $K$:
\begin{equation}
\mathbf{z}=\Phi_{\text{IAA}}(\mathbf{F}_v)\in\mathbb{R}^{T\times K}.
\end{equation}
We then apply softmax to obtain the score distribution and compute the frame-wise IAA score as the expectation over the K levels:
\begin{equation}
s_t=\frac{1}{K-1}\sum_{k=1}^{K} k\cdot p_{t,k}, \quad
p_{t,k}=\frac{\exp(z_{t,k})}{\sum_{j=1}^{K}\exp(z_{t,j})},
\end{equation}
where \(p_{t,k}\) denotes the predicted probability of the \(k\)-th aesthetic level for frame \(t\), and \(s_t\) is the resulting IAA score.

As existing VAA datasets do not provide frame-level aesthetic annotations, we pretrain \( \Phi_{\text{IAA}} \) on the AVA~\cite{ava} dataset and use it as a fixed perceptual prior in our framework. This enables frame-wise aesthetic knowledge learned from images to be transferred to VAA modeling without requiring additional annotations.

$\star$ FAP provides an explicit estimate of per-frame aesthetic quality, offering a basis for capturing the subtle interplay between individual frame quality and overall video aesthetics.

\textbf{Peak-End Rule-Based Temporal Aggregation.} Given frame-level visual features \( \mathbf{F}_v=\{\mathbf{f}_t\}_{t=1}^{T} \), where \( \mathbf{f}_t\in\mathbb{R}^{d} \), we aim to aggregate them into a holistic video representation. Rather than applying uniform temporal averaging, we perform non-uniform pooling inspired by the \textit{peak-end rule}~\cite{peak-end}, which states that human judgments of a temporal experience are dominated by a few salient moments, especially peaks and the ending.

We use the normalized frame-wise IAA scores \(\{s_t\}_{t=1}^{T}\), predicted by FAP, as a surrogate signal for temporal saliency. Since the video is represented by sampled frames, neighboring indices \(t\) and \(t+1\) represent two temporal regions separated by several intermediate raw frames, rather than two consecutive frames. This makes comparisons between adjacent sampled indices a plausible way to identify aesthetically salient moments.

$\triangleright$ \textit{Peak and Valley Discovery.} We first define \textit{peak} and \textit{valley}:

\begin{tcolorbox}[
    colback=gray!10,
    colframe=gray!10,
    boxrule=0pt,
    arc=3pt,
    left=8pt,
    right=8pt,
    top=5pt,
    bottom=5pt
]
\textbf{Definition 1 (Peak and Valley).}
For a sequence of sampled frame-level aesthetic scores \(\{s_t\}_{t=1}^{T}\), frame \(t\) is a \textit{peak} if \(s_t > s_{t-1}\) and \(s_t > s_{t+1}\), and a \textit{valley} if \(s_t < s_{t-1}\) and \(s_t < s_{t+1}\).
\end{tcolorbox}
To characterize such local extrema, we compute the temporal differences between each sampled frame and its neighbors as follows:
\begin{equation}
\begin{aligned}
& \Delta_t^{L+}=s_t-s_{t-1}, \quad \Delta_t^{R+}=s_t-s_{t+1},\\
& \Delta_t^{L-}=s_{t-1}-s_t, \quad \Delta_t^{R-}=s_{t+1}-s_t,
\end{aligned}
\end{equation}
where boundary values are set to zero when neighboring sampled frames are unavailable. Accordingly, the peak and valley responses at frame \(t\) are defined as:
\begin{equation}
\begin{aligned}
& r_t^{\text{peak}}=\mathrm{ReLU}(\Delta_t^{L+}) \cdot \mathrm{ReLU}(\Delta_t^{R+}) \cdot s_t,\\
& r_t^{\text{valley}}=\mathrm{ReLU}(\Delta_t^{L-}) \cdot \mathrm{ReLU}(\Delta_t^{R-}) \cdot (1-s_t).
\end{aligned}
\label{peakvalley}
\end{equation}
By construction, \textit{peak} and \textit{valley} are mutually exclusive. The product of the left and right difference terms amplifies local trends, thereby suppressing trivial fluctuations and highlighting clear local extrema. In addition, as valley frames usually correspond to low \(s_t\), the factor \(1-s_t\) is used to emphasize valley responses. These two terms form the basis for the computation of frame weights.

$\triangleright$ \textit{End Effect.} To account for the end effect of the \textit{peak-end rule}, we assign larger weights to frames closer to the end of the sequence. Specifically, we adopt an exponentially decaying function:
\begin{equation}
r^\text{end}_t=\exp(-\lambda d_t)\cdot s_t
\end{equation}
where \(d_t=\frac{T-t}{T}\) is the relative distance from frame \(t\) to the end, and \(\lambda=\frac{T}{K}\) controls the effective end span, with \(K\) being a hyperparameter specifying the number of emphasized ending frames.

$\triangleright$ \textit{Attention Weight Computation.} We then combine the three cues into an attention logit \(r_t\):
\begin{equation}
r_t=\sigma(r^{\text{base}})+\alpha\cdot r^\text{peak}_t+\beta\cdot r^\text{valley}_t+\gamma\cdot r^\text{end}_t,
\end{equation}
where \(r^{\text{base}} \in \mathbb{R}\) is a learnable base response applied to all frames, and \(\alpha, \beta, \gamma \in \mathbb{R}\) are also learnable coefficients that balance the contributions of different terms. The final video-level representation is computed as follows:
\begin{equation}
\mathbf{f}_{\text{video}}=\sum_{t=1}^{T} w_t \mathbf{f}_t\in\mathbb{R}^{d}, \quad
w_t=\frac{\exp(r_t)}{\sum_{j=1}^{T}\exp(r_j)}.
\end{equation}

$\star$ Compared with uniform averaging, this module explicitly highlights salient moments and emphasizes the ending, thereby providing an interpretable and psychologically grounded temporal inductive bias for video aesthetic assessment.

\textbf{Aesthetic Rhythm Pattern Encoding.} While peak-end aggregation captures the pivotal moments, it does not explicitly preserve the global evolution pattern. Yet videos with similar aggregated scores may still induce different impressions if their aesthetic trajectories evolve differently. We therefore first formalize the notion of \textit{aesthetic rhythm pattern} as shown below:

\begin{tcolorbox}[
    colback=gray!10,
    colframe=gray!10,
    boxrule=0pt,
    arc=3pt,
    left=8pt,
    right=8pt,
    top=5pt,
    bottom=5pt
]
\textbf{Definition 2 (Aesthetic Rhythm Pattern).}
For the frame-wise aesthetic score sequence, the aesthetic rhythm pattern refers to its temporal variation structure, including rising or falling tendency, local fluctuations, and overall progression over time. It characterizes how aesthetic quality evolves throughout the video, rather than merely its average level.
\end{tcolorbox}

To capture such patterns, we introduce a lightweight rhythm encoder, denoted by \(\Phi_{\text{RE}}(\cdot)\). It treats the sequence \(\mathbf{s}=[s_1,\ldots,s_T]\) as a one-dimensional aesthetic signal and extracts its rhythm patterns, thereby modeling how aesthetic quality evolves throughout the video:
\begin{equation}
\mathbf{f}_{\text{rhythm}}=\Phi_{\text{RE}}(\mathbf{s})\in\mathbb{R}^{d'}.
\end{equation}

$\star$ The representation encodes overall progression trends, thereby preserving the temporal ordering of aesthetic changes. It provides a complementary cue to peak-end aggregation.

\textbf{Training Objective.}
The model is trained with a video-level regression loss. Specifically, we concatenate \(\mathbf{f}_{\text{video}}\) and \(\mathbf{f}_{\text{rhythm}}\), and feed the fused representation into a learnable VAA head \(\Phi_{\text{VAA}}(\cdot)\) to predict the multi-dimensional aesthetic scores:
\begin{equation}
\hat{\mathbf{y}}=\sigma\!\left(\Phi_{\text{VAA}}\!\left(\mathbf{f}_{\text{video}}\oplus\mathbf{f}_{\text{rhythm}}\right)\right)\in\mathbb{R}^{11},
\end{equation}
where \(\oplus\) denotes feature concatenation, and \(\sigma(\cdot)\) is the sigmoid function. Following VADB~\cite{vadb}, the target vector \(\mathbf{y}\in\mathbb{R}^{11}\) includes 1 overall score, 6 general scores, and 4 character-specific scores. Since the character-specific dimensions are only valid for videos containing human subjects, we introduce a binary mask \(\mathbf{m}\in\{0,1\}^{11}\) to ignore unavailable annotations. The final loss is defined as:
\begin{equation}
\mathcal{L}_\text{VAA}
=\mathbb{E}_{(\mathbf{V},\mathbf{y})\in\mathcal{D}_{\text{VADB}}}
\left[
\frac{1}{\sum_{i=1}^{11} m_i}
\sum_{i=1}^{11} m_i\,(\hat{y}_i-y_i)^2
\right].
\end{equation}

\subsection{Adaptive Overall Enhancement}
\label{3.3}

Although the previous stage is supervised with fine-grained annotations, the overall-score prediction produced by the VAA head may still be unreliable in certain cases, and it may generalize poorly when encountering diverse video content.

$\triangleright$ \textit{Dynamic Gated Fusion.} To address the above limitations, we introduce a gated fusion module that adaptively integrates an image-aesthetic term into the final overall-score prediction. Specifically, given the frame-wise IAA scores \(\{s_t\}_{t=1}^{T}\), we obtain a global IAA score by \(\hat{y}_{\text{IAA}}=\frac{1}{T}\sum_{t=1}^{T} s_t\). Meanwhile, the VAA head outputs an 11-dimensional score vector
\(\hat{\mathbf{y}}=[\hat{y}_1,\ldots,\hat{y}_{11}]\), in which \(\hat{y}_1\) corresponds to the overall score, referred to as \(\hat{y}_\text{overall}\). We employ a gated fusion module \(\Phi_{\text{GF}}(\cdot)\) to adaptively balance these two estimates:
\begin{equation}
\begin{aligned}
& \lambda=\Phi_{\text{GF}}(\mathbf{f}_{\text{video}} \oplus \mathbf{f}_{\text{rhythm}})\in[0,1], \\
& \hat{y}^{\,\text{fusion}}_{\text{overall}}
=
\lambda\,\hat{y}_{\text{overall}}
+
(1-\lambda)\,\hat{y}_{\text{IAA}},
\end{aligned}
\label{gate}
\end{equation}
where \(\lambda\) controls the relative contribution of the VAA prediction and the global IAA score for each video.

To train the gate, we define a soft target according to the relative errors of the two estimates with respect to the ground truth:
\begin{equation}
\lambda^{*}
=\frac{|\hat{y}_{\text{IAA}}-y_{\text{overall}}|}
{|\hat{y}_{\text{IAA}}-y_{\text{overall}}|+|\hat{y}_{\text{overall}}-y_{\text{overall}}|+\epsilon},
\end{equation}
where \(\epsilon\) is a small constant for numerical stability. The training objective is then formalized as:
\begin{equation}
\begin{aligned}
\mathcal{L}_{GF}=\mathbb{E}_{(\mathbf{V},\mathbf{y})\in\mathcal{D}_{\text{VADB}}} 
\left[
(\hat{y}^{\,\text{fusion}}_{\text{overall}}-y_{\text{overall}})^2 +
\mathrm{BCE}(\lambda,\lambda^{*})\right].
\end{aligned}
\end{equation}

$\star$ By adaptively integrating the image-aesthetic term into overall-score prediction, this stage improves the robustness and generalizability of the overall video aesthetic assessment, especially when the direct prediction of the VAA head is less reliable.

\subsection{Inference Workflow}
\label{3.4}
Given an input video, we uniformly sample $T$ frames and extract frame-level visual features using the frozen ViT encoder. These features are fed into FAP to obtain frame-wise IAA scores $\{s_t\}_{t=1}^T$, which serve as aesthetic priors for the \textit{peak-end rule}-based temporal aggregation to compute a holistic video representation $\mathbf{f}_{\text{video}}$, while the score sequence is simultaneously encoded by the rhythm encoder to produce $\mathbf{f}_{\text{rhythm}}$. The concatenated representation $\mathbf{f}_{\text{video}} \oplus \mathbf{f}_{\text{rhythm}}$ is passed through the VAA head to predict an 11-dimensional score vector, including one overall score and ten attribute scores. For the final prediction, the overall score is refined via gated fusion (Eq.~\ref{gate}), which adaptively combines the VAA-predicted overall score with the global IAA score. The remaining attribute scores are directly taken from the VAA head output.

\textbf{Note:} the model components, including \(\Phi_{\text{IAA}}\), \(\Phi_{\text{VAA}}\), \(\Phi_{\text{RE}}\), and \(\Phi_{\text{GF}}\), are implemented using lightweight modules such as linear layers and 1D convolutions, with far fewer parameters than the backbone.

\section{Experiments}
\begin{table*}[t]
\centering
\caption{Performance comparison on the VADB test set across all eleven dimensions, evaluated using four metrics. Best results are in bold and second-best results are underlined. Our \textit{Peak-End-Net} is trained in a two-stage manner, where the two numbers reported in the \textit{Overall} column are presented as “Stage 1/Stage 2”.}
\label{tab:vadb_multidim_results}
\resizebox{\textwidth}{!}{
\begin{tabular}{l|l|ccccccccccc}
\toprule
\multirow{2}{*}{\textbf{Metric}} & \multirow{2}{*}{\textbf{Method}} & \multicolumn{11}{c}{\textbf{Evaluation on VADB~\cite{vadb} Across Different Evaluation Dimensions}} \\
& & \cellcolor{gray!7}\textbf{Overall} & \cellcolor{gray!7}\textbf{Com} & \cellcolor{gray!7}\textbf{SS} & \cellcolor{gray!7}\textbf{Lig} & \cellcolor{gray!7}\textbf{V\&T} & \cellcolor{gray!7}\textbf{Col} & \cellcolor{gray!7}\textbf{D\&F} & \cellcolor{gray!7}\textbf{Exp} & \cellcolor{gray!7}\textbf{Mov} & \cellcolor{gray!7}\textbf{Cos} & \cellcolor{gray!7}\textbf{Mak} \\
\midrule

\multirow{8}{*}{\textbf{RMSE}$\downarrow$}
& \cellcolor{blue!1}FastVQA~\cite{fastvqa}      & 0.6920 & 0.5929 & 0.6261 & 0.6710 & 0.7194 & 0.8220 & 0.7727 & 0.9948 & 0.9153 & 0.8459 & \underline{1.1922} \\
& \cellcolor{blue!1}SimpleVQA~\cite{simplevqa}    & 0.7538 & 0.6460 & 0.6811 & 0.7276 & 0.7723 & 0.8972 & 0.8457 & 1.0707 & 0.9936 & 0.9089 & 1.2639 \\
& \cellcolor{blue!1}ModularBVQA~\cite{modularbvqa}  & 0.6999 & 0.6147 & 0.6498 & 0.6875 & 0.7479 & 0.8702 & 0.8239 & 0.9488 & 0.9163 & 0.8181 & \textbf{1.1685} \\
& \cellcolor{blue!1}CLIPVQA~\cite{clipvqa}      & 0.6909 & 0.6140 & 0.6244 & 0.6669 & 0.7101 & 0.8328 & 0.7634 & 1.0316 & 0.9275 & 0.8646 & 1.2309 \\
& \cellcolor{blue!1}Q-Align~\cite{qalign}      & 0.6649 & 0.5616 & 0.5831 & 0.6320 & 0.6907 & 0.8173 & 0.7537 & 1.0070 & 0.9289 & 0.8500 & 1.2318 \\
& \cellcolor{blue!1}DOVER~\cite{divide}        & 0.7252 & 0.6282 & 0.6647 & 0.7101 & 0.7592 & 0.8917 & 0.8061 & 1.0295 & 0.9726 & 0.8870 & 1.2536 \\
& \cellcolor{blue!1}VADB-Net~\cite{vadb}         & 0.4998 & \underline{0.4067} & \underline{0.4421} & \underline{0.4877} & \underline{0.5589} & \textbf{0.7614} & \underline{0.6266} & \underline{0.8402} & \underline{0.6786} & \underline{0.6738} & 1.3751 \\
& \cellcolor{blue!7}\textbf{Peak-End-Net (Ours)}  & \textbf{0.4344}/\underline{0.4548} & \textbf{0.3180} & \textbf{0.3636} & \textbf{0.4172} & \textbf{0.4944} & \underline{0.7657} & \textbf{0.5776} & \textbf{0.6990} & \textbf{0.5417} & \textbf{0.5022} & 1.2075 \\
\midrule

\multirow{8}{*}{\textbf{SRCC}$\uparrow$}
& \cellcolor{blue!1}FastVQA~\cite{fastvqa}      & 0.8613 & 0.8737 & 0.8680 & 0.8536 & 0.8428 & 0.7873 & 0.8150 & 0.8273 & 0.8262 & 0.8298 & 0.7938 \\
& \cellcolor{blue!1}SimpleVQA~\cite{simplevqa}    & 0.8309 & 0.8565 & 0.8435 & 0.8207 & 0.8144 & 0.7469 & 0.7705 & 0.7921 & 0.7800 & 0.8023 & 0.7664 \\
& \cellcolor{blue!1}ModularBVQA~\cite{modularbvqa}  & 0.8549 & 0.8727 & 0.8573 & 0.8420 & 0.8268 & 0.7580 & 0.7842 & 0.8259 & 0.8098 & 0.8358 & 0.7809 \\
& \cellcolor{blue!1}CLIPVQA~\cite{clipvqa}      & 0.8543 & 0.8650 & 0.8656 & 0.8512 & 0.8456 & 0.7869 & 0.8137 & 0.8292 & 0.8395 & 0.8334 & 0.7984 \\
& \cellcolor{blue!1}Q-Align~\cite{qalign}      & \underline{0.8789} & \underline{0.8940} & \underline{0.8928} & \textbf{0.8757} & \underline{0.8609} & \underline{0.8004} & \underline{0.8297} & \underline{0.8438} & 0.8340 & \underline{0.8445} & 0.7925 \\
& \cellcolor{blue!1}DOVER~\cite{divide}        & 0.8606 & 0.8774 & 0.8684 & 0.8516 & 0.8423 & 0.7766 & 0.8155 & 0.8393 & 0.8369 & 0.8389 & 0.8089 \\
& \cellcolor{blue!1}VADB-Net~\cite{vadb}         & 0.8576 & 0.8726 & 0.8652 & 0.8525 & 0.8398 & 0.7729 & 0.8020 & 0.8428 & \underline{0.8502} & 0.8443 & \underline{0.7961} \\
& \cellcolor{blue!7}\textbf{Peak-End-Net (Ours)}  & \textbf{0.8875}/0.8784 & \textbf{0.8970} & \textbf{0.8976} & \underline{0.8755} & \textbf{0.8615} & \textbf{0.8038} & \textbf{0.8346} & \textbf{0.8646} & \textbf{0.8728} & \textbf{0.8726} & \textbf{0.8242} \\
\midrule

\multirow{8}{*}{\textbf{PLCC}$\uparrow$}
& \cellcolor{blue!1}FastVQA~\cite{fastvqa}      & 0.8577 & 0.9016 & 0.8897 & 0.8507 & 0.8450 & 0.7866 & 0.8063 & 0.8082 & 0.7939 & 0.8193 & 0.7551 \\
& \cellcolor{blue!1}SimpleVQA~\cite{simplevqa}    & 0.8285 & 0.8820 & 0.8680 & 0.8216 & 0.8187 & 0.7387 & 0.7622 & 0.7734 & 0.7513 & 0.7877 & 0.7190 \\
& \cellcolor{blue!1}ModularBVQA~\cite{modularbvqa}  & 0.8542 & 0.8938 & 0.8806 & 0.8425 & 0.8312 & 0.7567 & 0.7761 & 0.8273 & 0.7935 & \underline{0.8322} & \underline{0.7663} \\
& \cellcolor{blue!1}CLIPVQA~\cite{clipvqa}      & 0.8582 & 0.8941 & 0.8903 & 0.8526 & 0.8493 & 0.7801 & 0.8115 & 0.7918 & 0.7877 & 0.8103 & 0.7362 \\
& \cellcolor{blue!1}Q-Align~\cite{qalign}      & 0.8694 & \underline{0.9122} & \textbf{0.9050} & \underline{0.8688} & \underline{0.8580} & \underline{0.7892} & \underline{0.8167} & 0.8024 & 0.7869 & 0.8174 & 0.7359 \\
& \cellcolor{blue!1}DOVER~\cite{divide}        & 0.8549 & 0.8950 & 0.8825 & 0.8453 & 0.8407 & 0.7756 & 0.8097 & 0.8143 & 0.7913 & 0.8193 & 0.7624 \\
& \cellcolor{blue!1}VADB-Net~\cite{vadb}         & 0.8546 & 0.8871 & 0.8778 & 0.8479 & 0.8384 & 0.7700 & 0.7989 & \underline{0.8316} & \underline{0.8274} & 0.8307 & 0.7631 \\
& \cellcolor{blue!7}\textbf{Peak-End-Net (Ours)}  & \textbf{0.8853}/\underline{0.8765} & \textbf{0.9128} & \underline{0.9013} & \textbf{0.8717} & \textbf{0.8636} & \textbf{0.7915} & \textbf{0.8272} & \textbf{0.8664} & \textbf{0.8718} & \textbf{0.8735} & \textbf{0.7983} \\
\midrule

\multirow{8}{*}{\textbf{KRCC}$\uparrow$}
& \cellcolor{blue!1}FastVQA~\cite{fastvqa}      & 0.6647 & 0.6756 & 0.6697 & 0.6569 & 0.6401 & 0.5847 & 0.6169 & 0.6385 & 0.6359 & 0.6378 & 0.5986 \\
&\cellcolor{blue!1}SimpleVQA~\cite{simplevqa}    & 0.6316 & 0.6535 & 0.6389 & 0.6181 & 0.6103 & 0.5478 & 0.5703 & 0.5982 & 0.5846 & 0.6074 & 0.5705 \\
& \cellcolor{blue!1}ModularBVQA~\cite{modularbvqa}  & 0.6604 & 0.6751 & 0.6572 & 0.6415 & 0.6266 & 0.5576 & 0.5861 & 0.6383 & 0.6173 & 0.6447 & 0.5888 \\
& \cellcolor{blue!1}CLIPVQA~\cite{clipvqa}      & 0.6515 & 0.6630 & 0.6678 & 0.6496 & 0.6418 & 0.5783 & 0.6074 & 0.6406 & 0.6493 & 0.6395 & 0.5992 \\
& \cellcolor{blue!1}Q-Align~\cite{qalign}      & 0.6903 & \underline{0.7037} & \underline{0.7054} & \textbf{0.6853} & \underline{0.6669} & \underline{0.6022} & \underline{0.6392} & \underline{0.6624} & 0.6489 & \underline{0.6607} & 0.6078 \\
& \cellcolor{blue!1}DOVER~\cite{divide}        & 0.6636 & 0.6802 & 0.6705 & 0.6523 & 0.6408 & 0.5756 & 0.6174 & 0.6516 & 0.6460 & 0.6479 & \underline{0.6168} \\
& \cellcolor{blue!1}VADB-Net~\cite{vadb}         & 0.6587 & 0.6725 & 0.6638 & 0.6549 & 0.6421 & 0.5747 & 0.6050 & 0.6504 & \underline{0.6520} & 0.6522 & 0.6022 \\
& \cellcolor{blue!7}\textbf{Peak-End-Net (Ours)}  & \textbf{0.7014}/\underline{0.6925} & \textbf{0.7113} & \textbf{0.7062} & \underline{0.6833} & \textbf{0.6751} & \textbf{0.6083} & \textbf{0.6447} & \textbf{0.6811} & \textbf{0.6878} & \textbf{0.6919} & \textbf{0.6373} \\
\bottomrule
\end{tabular}
}
\end{table*}
\subsection{Experimental Settings}
\textbf{Evaluation Benchmarks.}
We evaluate our method on VADB~\cite{vadb} and DIVIDE-3K~\cite{divide}. \ding{182} VADB is used for training and in-domain evaluation. It contains short videos from social media platforms, professional cinematography, and movie clips, providing one overall score, six general-dimension scores, and four character-specific-dimension scores, all annotated by professionals from the Beijing Film Academy. Due to copyright restrictions, only 7,881 videos are publicly available. Therefore, all experiments on VADB are conducted on this subset, which we randomly split into training and test sets with a ratio of 8:2. Results are reported on the test split for both the overall score and other annotated dimensions. \ding{183} We further conduct cross-dataset testing on DIVIDE-3K, a benchmark containing 3,590 in-the-wild videos with disentangled overall, aesthetic, and technical annotations. Models trained on VADB are directly applied to DIVIDE-3K without fine-tuning. Since DIVIDE-3K provides only one aesthetic-related annotation per video in our setting, we report zero-shot performance on this target dataset.

\textbf{Evaluation Metrics.} Following prior works~\cite{vadb,divide,fastvqa,simplevqa,clipvqa}, we adopt four standard evaluation metrics: Root Mean Square Error (RMSE), Spearman Rank Correlation Coefficient (SRCC), Pearson Linear Correlation Coefficient (PLCC), and Kendall Rank Correlation Coefficient (KRCC). RMSE measures absolute prediction error, where lower values are better. SRCC and KRCC assess ranking consistency, while PLCC measures linear correlation; for these three metrics, higher values indicate better performance. On VADB~\cite{vadb}, we report all four metrics for the overall score and each annotated attribute. On DIVIDE-3K~\cite{divide}, we report the same metrics on its aesthetic dimension under the zero-shot evaluation protocol.

\textbf{Implementation Details.} All experiments are conducted on 8 NVIDIA A6000 GPUs to accelerate training, although our method can also be trained on a single GPU with just 16 GB GPU memory. We adopt ViT-L/14~\cite{vit}, initialized with OpenAI CLIP~\cite{clip} pretrained weights, as the frozen visual backbone to extract frame-level representations. Unless otherwise specified, 12 frames are uniformly sampled from each video by default during both training and evaluation, and all input frames are resized to \(224\times224\). The training procedure consists of two stages. In both stages, we use AdamW~\cite{adamw} as the optimizer, with a batch size of 128 and a learning rate of \(1\times10^{-3}\). Stage 1 is trained for 10 epochs with 3 warm-up epochs, while Stage 2 is trained for 15 epochs with 2 warm-up epochs. For modeling the end effect, we set the hyperparameter \(K=4\), which means that the last one-third of the sampled frames are emphasized.

\subsection{Quantitative Results}
\begin{figure}[t]
\centering
\includegraphics[width=\columnwidth]{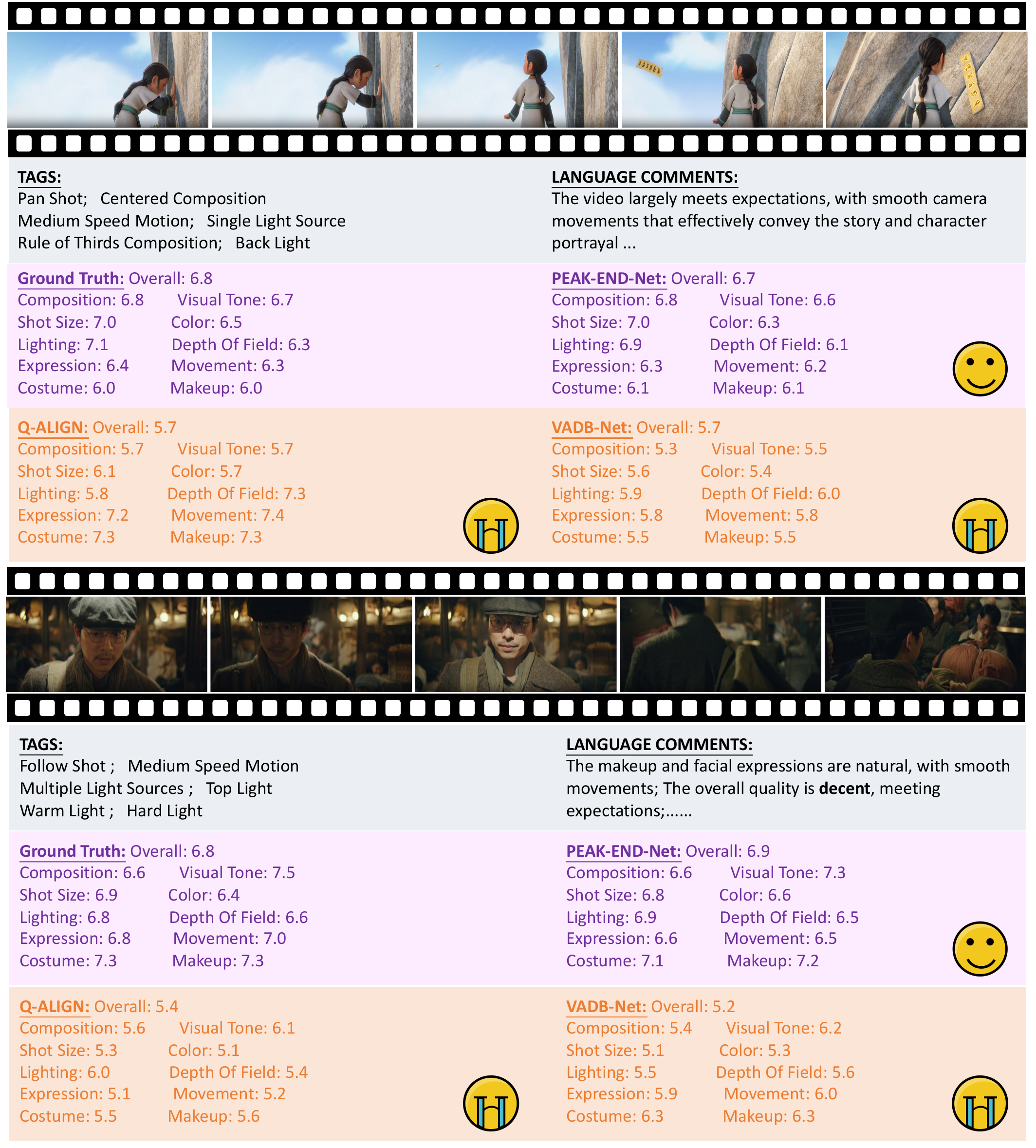}
\caption{Qualitative results on two VADB examples. We present the tags, language comments, ground-truth annotations, and the predicted overall and attribute-level scores from \textit{Peak-End-Net}, \textit{Q-Align}, and \textit{VADB-Net}.}
\Description{Qualitative comparison on two VADB examples showing tags, comments, ground-truth scores, and predictions from three methods.}
\label{fig:sample}
\end{figure}

\textbf{In-Domain Evaluation on VADB.} Table~\ref{tab:vadb_multidim_results} presents the in-domain results on VADB. Our \textit{Peak-End-Net} achieves the best overall performance, ranking first on the overall score under all four metrics, with \textbf{0.4344} on RMSE, \textbf{0.8875} on SRCC, \textbf{0.8853} on PLCC, and \textbf{0.7014} on KRCC. It also performs consistently well across fine-grained attributes, achieving the best results on most dimensions under SRCC, PLCC, and KRCC, as well as the lowest RMSE. Notably, \textit{Peak-End-Net} outperforms \textit{Q-Align}, a 7B-parameter LLM-based model, while relying only on a much smaller frozen ViT backbone, demonstrating the effectiveness and efficiency of our approach.

\begin{table}[t]
\centering
\caption{Zero-shot performance on the DIVIDE-3K test set. Best results are marked in bold. We separately present the results of the two-stage \textit{Peak-End-Net}.}
\label{tab:divide3k_zeroshot}
\resizebox{\columnwidth}{!}{
\begin{tabular}{l|cccc}
\toprule
\multirow{2}{*}{\textbf{Method}} & \multicolumn{4}{c}{\textbf{DIVIDE-3K~\cite{divide} Aesthetic Dimension}} \\
& \cellcolor{gray!7}\textbf{RMSE}$\downarrow$ & \cellcolor{gray!7}\textbf{SRCC}$\uparrow$ & \cellcolor{gray!7}\textbf{PLCC}$\uparrow$ & \cellcolor{gray!7}\textbf{KRCC}$\uparrow$ \\
\midrule
\cellcolor{blue!1}FastVQA~\cite{fastvqa} & 0.5826     & 0.1054      & 0.1442     & 0.0713     \\
\cellcolor{blue!1}SimpleVQA~\cite{simplevqa} & 0.5515 & 0.3279 & 0.3502 & 0.2252 \\
\cellcolor{blue!1}ModularBVQA~\cite{modularbvqa} & 0.5686 & 0.2437 & 0.2597 & 0.1642 \\
\cellcolor{blue!1}DOVER~\cite{divide} & 0.6282 & 0.3998 & 0.4308 & 0.2747 \\
\cellcolor{blue!1}CLIPVQA~\cite{clipvqa} & 0.5581 & 0.3035     & 0.3184    & 0.2027       \\
\cellcolor{blue!1}Q-Align~\cite{qalign} & 0.5151 & 0.4695 & 0.4842 & 0.3266 \\
\cellcolor{blue!1}VADB-Net~\cite{vadb} & 0.5772 & 0.1799 & 0.1968 & 0.1228 \\
\cellcolor{blue!7}\textbf{Peak-End-Net (Stage 1)} & 0.5155 & 0.4600 & 0.4832 & 0.3009 \\
\cellcolor{blue!7}\textbf{Peak-End-Net (Stage 2)} & \textbf{0.4905} & \textbf{0.5427} & \textbf{0.5773} & \textbf{0.3745} \\
\bottomrule
\end{tabular}
}
\end{table}

We observe that Stage 2 leads to a slight drop in in-domain performance on VADB. This is an intentional trade-off, as Stage 2 is designed to improve cross-dataset transferability and zero-shot generalization, while mitigating overfitting to the training data, rather than to further improve in-domain performance. Despite this small decrease, the resulting performance is highly competitive and remains the state-of-the-art.

Overall, these results validate the effectiveness of our design for multi-dimensional video aesthetic assessment. In particular, the \textit{peak-end rule}-based aggregation introduces an explicit temporal inductive bias that emphasizes aesthetically salient moments and the ending segment of a video, while the transferred image aesthetic prior provides useful frame-level guidance in the absence of dense frame-wise annotations.

\textbf{Cross-Domain Transferability on DIVIDE-3K.} Table~\ref{tab:divide3k_zeroshot} reports the zero-shot transfer results from VADB to DIVIDE-3K. Under this challenging setting, \textit{Peak-End-Net (Stage 2)} achieves the best performance on all four metrics, with \textbf{0.4905} on RMSE, \textbf{0.5427} on SRCC, \textbf{0.5773} on PLCC, and \textbf{0.3745} on KRCC, consistently outperforming all competing methods. More importantly, the comparison between Stage 1 and Stage 2 directly validates the effectiveness of the proposed \textit{Dynamic Gated Fusion}. Although Stage 1 is already competitive, Stage 2 brings consistent improvements across all four metrics, indicating that the second stage is effective in enhancing transferability. This gain can be attributed to the design of Stage 2. Specifically, the overall prediction from the VAA head is learned on VADB and may overfit to the training data or inherit source-domain bias, whereas the IAA prior provides a lower-level and more transferable aesthetic signal derived from large-scale image-level supervision. By fusing these two sources through a gating mechanism, Stage 2 produces more robust overall predictions.

Overall, the zero-shot results demonstrate that the proposed second stage is crucial for generalization. They confirm that image-aesthetic priors are valuable not only for frame-level guidance, but also for improving the robustness of overall aesthetic assessment in cross-dataset evaluation.

\begin{table}[t]
\centering
\caption{Ablation study of temporal aggregation and model components on the VADB test set. For brevity, we report only the results on the \textit{Overall} dimension. The best results are marked in bold.}
\resizebox{\columnwidth}{!}{
\begin{tabular}{l|cccc}
\toprule
\multirow{2}{*}{\textbf{Variant}} & \multicolumn{4}{c}{\textbf{VADB~\cite{vadb} Overall Dimension}} \\
& \cellcolor{gray!7}\textbf{RMSE}$\downarrow$ & \cellcolor{gray!7}\textbf{SRCC}$\uparrow$ & \cellcolor{gray!7}\textbf{PLCC}$\uparrow$ & \cellcolor{gray!7}\textbf{KRCC}$\uparrow$ \\
\midrule
\multicolumn{5}{@{}c@{}}{\textit{Comparison with alternative temporal aggregation strategies.}} \\
\addlinespace[2pt]
\cellcolor{blue!1}Mean Pooling & 0.4771 & 0.8680 & 0.8666 & 0.6715 \\
\cellcolor{blue!1}Max Pooling & 0.4768 & 0.8711 & 0.8690 & 0.6747 \\
\cellcolor{blue!1}Temporal Transformer & 0.4654 & 0.8683 & 0.8710 & 0.6742 \\
\midrule
\multicolumn{5}{@{}c@{}}{\textit{Different variants of our proposed method}} \\
\addlinespace[2pt]
\cellcolor{blue!7}Peak-End + \(\Phi_\text{random}\) & 0.4548 & 0.8804 & 0.8785 & 0.6925 \\
\cellcolor{blue!7}Peak-End + \(\Phi_\text{IAA}\) & 0.4355 & 0.8857 & 0.8832 & 0.6939 \\
\cellcolor{blue!7}Peak-End + \(\Phi_\text{IAA}\) + \(\Phi_\text{RE}\) & \textbf{0.4344} & \textbf{0.8875} & \textbf{0.8853} & \textbf{0.7014} \\
\bottomrule
\end{tabular}
}
\label{tab:method_comparison}
\end{table}

\begin{table}[t]
\centering
\caption{Ablation study of different \textit{peak-end} aggregation components on the VADB test set. Best results are in bold.}
\resizebox{\columnwidth}{!}{
\begin{tabular}{ccc|cccc}
\toprule
\multirow{2}{*}{\textbf{Peak}} & \multirow{2}{*}{\textbf{End}} & \multirow{2}{*}{\textbf{Valley}} & \multicolumn{4}{c}{\textbf{VADB~\cite{vadb} Overall Dimension}} \\
& & & \cellcolor{gray!7}\textbf{RMSE}$\downarrow$ & \cellcolor{gray!7}\textbf{SRCC}$\uparrow$ & \cellcolor{gray!7}\textbf{PLCC}$\uparrow$ & \cellcolor{gray!7}\textbf{KRCC}$\uparrow$ \\
\midrule
\cmark & \xmark & \xmark & 0.4802 & 0.8688 & 0.8677 & 0.6805 \\
\xmark & \cmark & \xmark & 0.4885 & 0.8693 & 0.8683 & 0.6743 \\
\xmark & \xmark & \cmark & 0.4954 & 0.8684 & 0.8677 & 0.6732 \\
\cmark & \cmark & \xmark & 0.4842 & 0.8739 & 0.8726 & 0.6803 \\
\cmark & \xmark & \cmark & 0.4957 & 0.8737 & 0.8727 & 0.6798 \\
\xmark & \cmark & \cmark & 0.4802 & 0.8741 & 0.8727 & 0.6805 \\
\cmark & \cmark & \cmark & \textbf{0.4548} & \textbf{0.8804} & \textbf{0.8785} & \textbf{0.6925} \\
\bottomrule
\end{tabular}
}
\label{tab:peak_end_ablation}
\end{table}

\subsection{Qualitative Results}
Figure~\ref{fig:sample} presents two representative examples from VADB. In both cases, \textit{Peak-End-Net} produces predictions that are much closer to the ground truth than \textit{Q-Align} and \textit{VADB-Net}, not only on the overall score but also across fine-grained attribute dimensions. In the first example, \textit{Peak-End-Net} accurately captures the balanced aesthetic quality reflected in composition, shot size, and lighting, while the other two methods exhibit obvious deviations. In the second example, \textit{Peak-End-Net} again better matches the ground-truth annotations, especially on visual tone, movement, costume, and makeup. These cases suggest that \textit{Peak-End-Net} is able to provide more reliable and consistent multi-dimensional aesthetic predictions.

\subsection{Ablation Study}

\textbf{Effectiveness of Peak-End Aggregation.} Table~\ref{tab:method_comparison} reports the ablation results on the \textit{Overall} dimension of VADB. \ding{182} \textit{Comparison with alternative temporal aggregation strategies.} We compare the proposed peak-end aggregation with mean pooling, max pooling, and a temporal transformer, where the latter aggregates frame features with a 4-layer, 8-head transformer encoder over a learnable \text{[CLS]} token. Peak-end aggregation outperforms all alternatives, improving RMSE from 0.4654 to 0.4548 and SRCC from 0.8683 to 0.8804 over the temporal transformer baseline. This shows that explicitly emphasizing salient moments and the ending segment is more effective than generic temporal aggregation for video aesthetic assessment. \ding{183} \textit{Effectiveness of the pretrained IAA head and rhythm encoder.} Replacing the randomly initialized IAA head $\Phi_\text{IAA}^{\text{random}}$ with the AVA-pretrained $\Phi_\text{IAA}$ consistently improves all four metrics, verifying the benefit of transferring image-aesthetic priors. Adding the rhythm encoder $\Phi_\text{RE}$ brings additional gains, leading to the best overall performance. This suggests that, in addition to emphasizing salient moments and the ending effect, modeling aesthetic progression is also beneficial for VAA.

\textbf{Components in Peak-End Aggregation.} Table~\ref{tab:peak_end_ablation} presents the ablation study of different components in the proposed \textit{peak-end rule}-based aggregation. To reduce the interference of other modules, we conduct the ablation study based on \textit{Peak-End + \(\Phi_\text{random}\)} in Table~\ref{tab:method_comparison}. \ding{182} \textit{Single-component variants.} Using only one component, i.e., \textit{Peak}, \textit{End}, or \textit{Valley}, yields limited performance, indicating that no single temporal cue is sufficient to capture holistic video aesthetics. Among them, the \textit{End}-only variant shows slightly better overall behavior, suggesting that recency information is an important factor in aesthetic judgment. \ding{183} \textit{Two-component variants.} Combining any two components generally leads to better results than using a single one, which suggests that these temporal cues are complementary and exhibit a certain degree of collaboration. This confirms that VAA benefits from integrating multiple temporal factors rather than relying on an isolated cue. \ding{184} \textit{Full peak-end aggregation.} The full model that jointly incorporates \textit{Peak}, \textit{End}, and \textit{Valley} achieves the best performance among all variants. This verifies that a more complete temporal representation, covering salient moments, ending effects, and unfavorable valleys, is most effective for VAA.
\begin{figure}[t]
\centering
\subfigure[End effect hyperparameter $K$ in the \textit{peak-end rule}-based aggregation.]{
\includegraphics[width=\columnwidth]{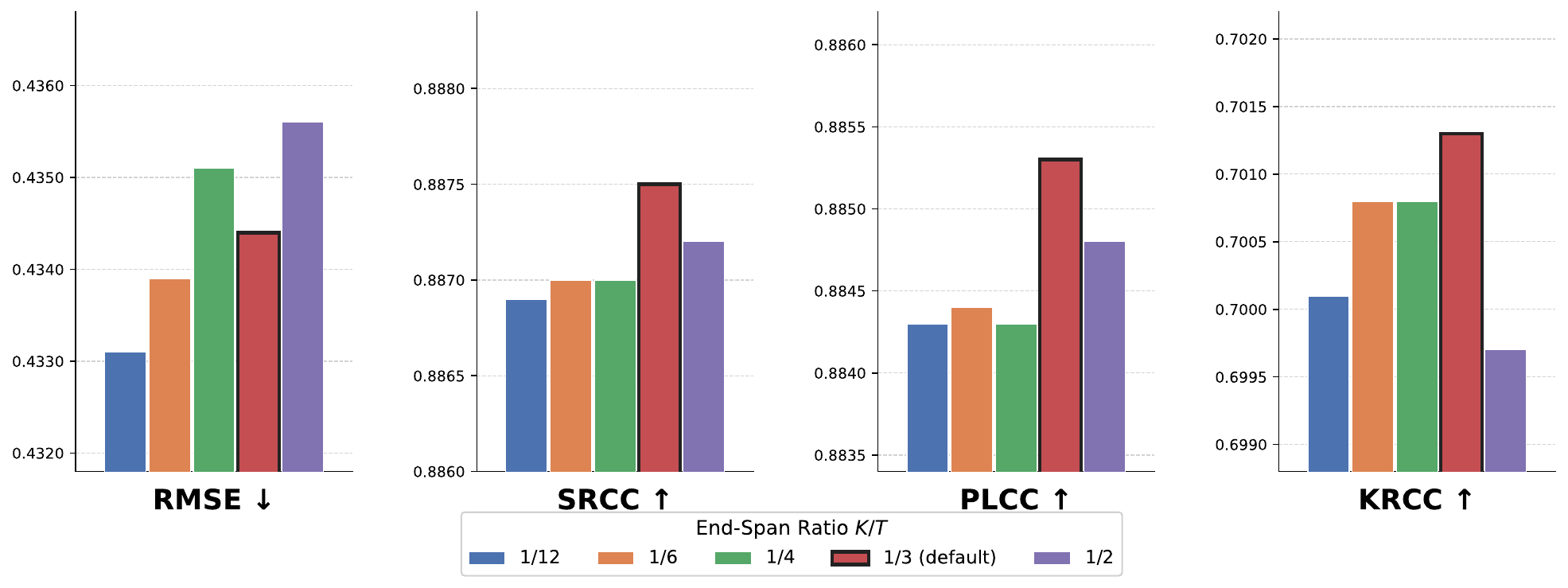}
\label{fig:kt}
}
\subfigure[Impact of frame sampling density.]{
\includegraphics[width=\columnwidth]{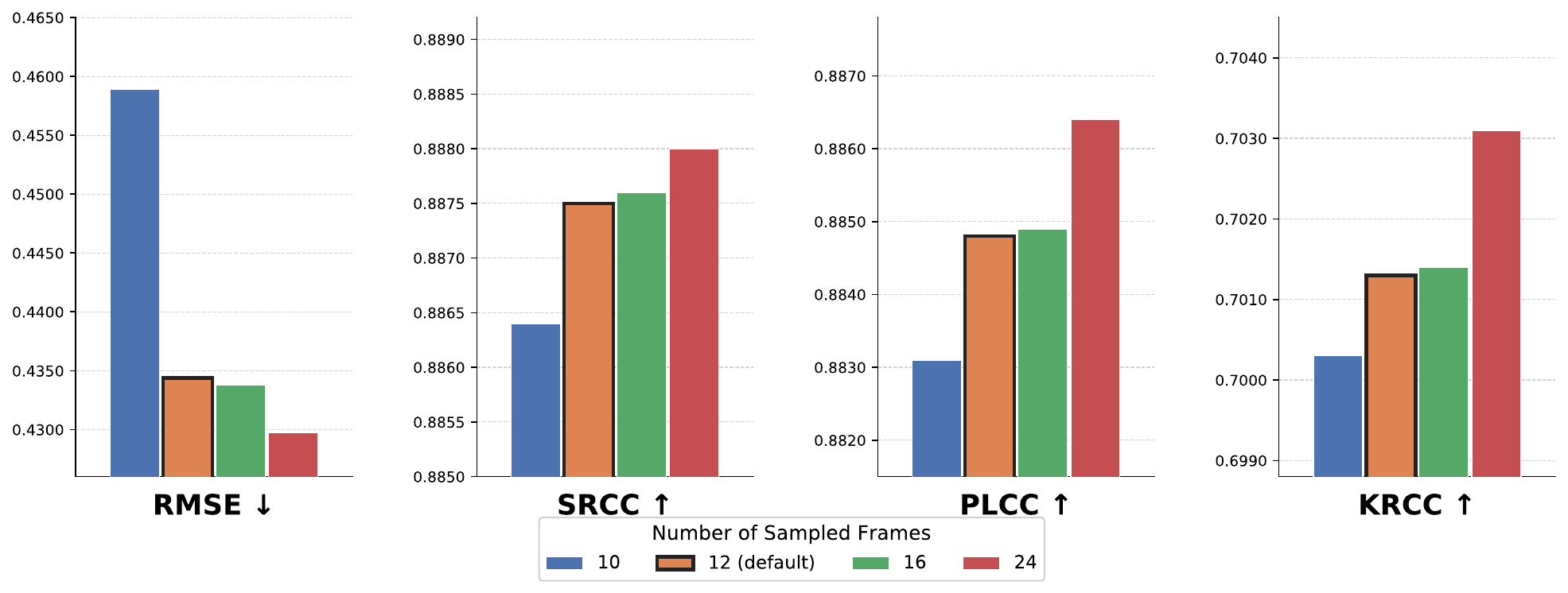}
\label{fig:frame}
}
\caption{Sensitivity analysis of key hyperparameters.}
\Description{Bar charts of four metrics under varying end-span ratio and varying number of sampled frames.}
\label{fig:sensitivity}
\end{figure}
\textbf{Sensitivity Analysis of Key Hyperparameters.} Figure~\ref{fig:sensitivity} analyzes two key hyperparameters. \ding{182} \textit{Effect of the end-span ratio \(K/T\).} With 12 sampled frames, using the last one-third of the video (\(K=4\)) gives the best performance, showing the importance of properly modeling the ending segment. \ding{183} \textit{Effect of the number of sampled frames.} Performance generally improves with more sampled frames, suggesting that denser temporal observation is beneficial. To balance accuracy and efficiency, we adopt 12 frames by default.

\begin{table}[t]
\centering
\caption{Ablation study of Stage 2 fusion strategies on the in-domain VADB test set and the cross-dataset DIVIDE-3K evaluation. Numbers in parentheses indicate the performance change relative to Stage 1. Best results are in bold.}
\resizebox{\columnwidth}{!}{
\begin{tabular}{c|l|ccc}
\toprule
\textbf{Dataset} & \textbf{Variant} & \cellcolor{gray!7}\textbf{SRCC}$\uparrow$ & \cellcolor{gray!7}\textbf{PLCC}$\uparrow$ & \cellcolor{gray!7}\textbf{KRCC}$\uparrow$ \\
\midrule
\multirow{4}{*}{\rotatebox[origin=c]{90}{\textbf{DIVIDE}}}
& \cellcolor{blue!1}Peak-End-Net (Stage 1)                & 0.4600 & 0.4832 & 0.3009 \\
& \cellcolor{blue!1}+ Static Average Fusion                 & 0.5285 {\footnotesize(+0.0685)} & 0.5746 {\footnotesize(+0.0914)} & 0.3860 {\footnotesize\textbf{(+0.0851)}} \\
& \cellcolor{blue!1}+ Learnable Scalar Fusion               & 0.5248 {\footnotesize(+0.0648)} & 0.5588 {\footnotesize(+0.0756)} & 0.3701 {\footnotesize(+0.0692)} \\
& \cellcolor{blue!7}+ \textbf{Dynamic Gated Fusion (Ours)}   & 0.5427 {\footnotesize\textbf{(+0.0827)}} & 0.5773 {\footnotesize\textbf{(+0.0941)}} & 0.3745 {\footnotesize(+0.0736)} \\
\midrule
\multirow{4}{*}{\rotatebox[origin=c]{90}{\textbf{VADB}}}
& \cellcolor{blue!1}Peak-End-Net (Stage 1)                & 0.8875 & 0.8853 & 0.7014 \\
& \cellcolor{blue!1}+ Static Average Fusion                 & 0.8353 {\footnotesize(-0.0522)} & 0.8391 {\footnotesize(-0.0462)} & 0.6334 {\footnotesize(-0.0679)} \\
& \cellcolor{blue!1}+ Learnable Scalar Fusion               & 0.8691 {\footnotesize(-0.0184)} & 0.8690 {\footnotesize(-0.0163)} & 0.6780 {\footnotesize(-0.0233)} \\
& \cellcolor{blue!7}+ \textbf{Dynamic Gated Fusion (Ours)}   & 0.8784 {\footnotesize\textbf{(-0.0091)}} & 0.8765 {\footnotesize\textbf{(-0.0088)}} & 0.6925 {\footnotesize\textbf{(-0.0089)}} \\
\bottomrule
\end{tabular}
}
\label{tab:stage2_ablation}
\end{table}

\textbf{Effectiveness of Dynamic Gated Fusion.} Table~\ref{tab:stage2_ablation} reports the comparison of different Stage 2 fusion strategies on both VADB and DIVIDE-3K. In this part, we further compare our method with two fusion baselines. \textit{Static Average Fusion} directly averages the Stage 1 overall prediction and the global IAA score with equal weights. \textit{Learnable Scalar Fusion} replaces the fixed averaging weight with a single learnable scalar shared by all samples. All fusion variants improve the cross-dataset performance on DIVIDE-3K over Stage 1, indicating that the global IAA estimate provides meaningful transferable information. \textit{Static Average Fusion} achieves the largest gain in KRCC, while the proposed \textit{Dynamic Gated Fusion} achieves the largest gains in SRCC and PLCC. 

The key difference lies in their impact on the source-domain VADB test set. 
\textit{Static Average Fusion} obtains cross-dataset gains at the cost of a substantial degradation on VADB, with clear drops in all three metrics. This suggests that its improvement mainly comes from sacrificing source-domain performance. 
\textit{Learnable Scalar Fusion} alleviates this issue to some extent, but still causes noticeable performance loss. In contrast, our dynamic gated fusion improves the cross-dataset performance while introducing only very small decreases on VADB. Therefore, our proposed method achieves a much more favorable trade-off between cross-dataset robustness and in-domain prediction quality.

\section{Conclusion}

In this paper, we presented \textit{Peak-End-Net}, a psychologically grounded framework 
for video aesthetic assessment inspired by the \textit{peak-end rule}. 
By leveraging frame-wise aesthetic priors from a pretrained IAA head, our method 
transfers image-level aesthetic knowledge to video-level modeling 
without requiring frame-level video annotations. 
Complemented by an aesthetic rhythm encoder that captures temporal progression and 
a dynamic gated fusion mechanism that adaptively integrates image-aesthetic priors 
for robust overall-score prediction, \textit{Peak-End-Net} achieves state-of-the-art 
performance on both settings: in-domain evaluation on VADB and zero-shot transfer to DIVIDE-3K.

\bibliographystyle{ACM-Reference-Format}
\bibliography{sample-base}


\begin{thebibliography}{49}


\ifx \showCODEN    \undefined \def \showCODEN     #1{\unskip}     \fi
\ifx \showISBNx    \undefined \def \showISBNx     #1{\unskip}     \fi
\ifx \showISBNxiii \undefined \def \showISBNxiii  #1{\unskip}     \fi
\ifx \showISSN     \undefined \def \showISSN      #1{\unskip}     \fi
\ifx \showLCCN     \undefined \def \showLCCN      #1{\unskip}     \fi
\ifx \shownote     \undefined \def \shownote      #1{#1}          \fi
\ifx \showarticletitle \undefined \def \showarticletitle #1{#1}   \fi
\ifx \showURL      \undefined \def \showURL       {\relax}        \fi
\providecommand\bibfield[2]{#2}
\providecommand\bibinfo[2]{#2}
\providecommand\natexlab[1]{#1}
\providecommand\showeprint[2][]{arXiv:#2}

\bibitem[Achiam et~al\mbox{.}(2023)]%
        {gpt4}
\bibfield{author}{\bibinfo{person}{Josh Achiam}, \bibinfo{person}{Steven
  Adler}, \bibinfo{person}{Sandhini Agarwal}, \bibinfo{person}{Lama Ahmad},
  \bibinfo{person}{Ilge Akkaya}, \bibinfo{person}{Florencia~Leoni Aleman},
  \bibinfo{person}{Diogo Almeida}, \bibinfo{person}{Janko Altenschmidt},
  \bibinfo{person}{Sam Altman}, \bibinfo{person}{Shyamal Anadkat},
  {et~al\mbox{.}}} \bibinfo{year}{2023}\natexlab{}.
\newblock \showarticletitle{Gpt-4 technical report}.
\newblock \bibinfo{journal}{\emph{arXiv preprint arXiv:2303.08774}}
  (\bibinfo{year}{2023}).
\newblock


\bibitem[Bai et~al\mbox{.}(2025)]%
        {qwen3vl}
\bibfield{author}{\bibinfo{person}{Shuai Bai}, \bibinfo{person}{Yuxuan Cai},
  \bibinfo{person}{Ruizhe Chen}, \bibinfo{person}{Keqin Chen},
  \bibinfo{person}{Xionghui Chen}, \bibinfo{person}{Zesen Cheng},
  \bibinfo{person}{Lianghao Deng}, \bibinfo{person}{Wei Ding},
  \bibinfo{person}{Chang Gao}, \bibinfo{person}{Chunjiang Ge}, {et~al\mbox{.}}}
  \bibinfo{year}{2025}\natexlab{}.
\newblock \showarticletitle{Qwen3-vl technical report}.
\newblock \bibinfo{journal}{\emph{arXiv preprint arXiv:2511.21631}}
  (\bibinfo{year}{2025}).
\newblock


\bibitem[Chen et~al\mbox{.}(2025)]%
        {chen2025finger}
\bibfield{author}{\bibinfo{person}{Rui Chen}, \bibinfo{person}{Lei Sun},
  \bibinfo{person}{Jing Tang}, \bibinfo{person}{Geng Li}, {and}
  \bibinfo{person}{Xiangxiang Chu}.} \bibinfo{year}{2025}\natexlab{}.
\newblock \showarticletitle{Finger: Content aware fine-grained evaluation with
  reasoning for ai-generated videos}. In \bibinfo{booktitle}{\emph{Proceedings
  of the 33rd ACM International Conference on Multimedia}}.
  \bibinfo{pages}{3517--3526}.
\newblock


\bibitem[Chen et~al\mbox{.}(2020)]%
        {simclr}
\bibfield{author}{\bibinfo{person}{Ting Chen}, \bibinfo{person}{Simon
  Kornblith}, \bibinfo{person}{Mohammad Norouzi}, {and}
  \bibinfo{person}{Geoffrey Hinton}.} \bibinfo{year}{2020}\natexlab{}.
\newblock \showarticletitle{A simple framework for contrastive learning of
  visual representations}. In \bibinfo{booktitle}{\emph{International
  conference on machine learning}}. PmLR, \bibinfo{pages}{1597--1607}.
\newblock


\bibitem[Chen et~al\mbox{.}(2024)]%
        {internvl}
\bibfield{author}{\bibinfo{person}{Zhe Chen}, \bibinfo{person}{Jiannan Wu},
  \bibinfo{person}{Wenhai Wang}, \bibinfo{person}{Weijie Su},
  \bibinfo{person}{Guo Chen}, \bibinfo{person}{Sen Xing},
  \bibinfo{person}{Muyan Zhong}, \bibinfo{person}{Qinglong Zhang},
  \bibinfo{person}{Xizhou Zhu}, \bibinfo{person}{Lewei Lu}, {et~al\mbox{.}}}
  \bibinfo{year}{2024}\natexlab{}.
\newblock \showarticletitle{Internvl: Scaling up vision foundation models and
  aligning for generic visual-linguistic tasks}. In
  \bibinfo{booktitle}{\emph{Proceedings of the IEEE/CVF conference on computer
  vision and pattern recognition}}. \bibinfo{pages}{24185--24198}.
\newblock


\bibitem[Datta et~al\mbox{.}(2006)]%
        {iaaphoto}
\bibfield{author}{\bibinfo{person}{Ritendra Datta}, \bibinfo{person}{Dhiraj
  Joshi}, \bibinfo{person}{Jia Li}, {and} \bibinfo{person}{James~Z Wang}.}
  \bibinfo{year}{2006}\natexlab{}.
\newblock \showarticletitle{Studying aesthetics in photographic images using a
  computational approach}. In \bibinfo{booktitle}{\emph{European conference on
  computer vision}}. Springer, \bibinfo{pages}{288--301}.
\newblock


\bibitem[Deng et~al\mbox{.}(2017)]%
        {iaa}
\bibfield{author}{\bibinfo{person}{Yubin Deng}, \bibinfo{person}{Chen~Change
  Loy}, {and} \bibinfo{person}{Xiaoou Tang}.} \bibinfo{year}{2017}\natexlab{}.
\newblock \showarticletitle{Image aesthetic assessment: An experimental
  survey}.
\newblock \bibinfo{journal}{\emph{IEEE Signal Processing Magazine}}
  \bibinfo{volume}{34}, \bibinfo{number}{4} (\bibinfo{year}{2017}),
  \bibinfo{pages}{80--106}.
\newblock


\bibitem[Dosovitskiy et~al\mbox{.}(2020)]%
        {vit}
\bibfield{author}{\bibinfo{person}{Alexey Dosovitskiy}, \bibinfo{person}{Lucas
  Beyer}, \bibinfo{person}{Alexander Kolesnikov}, \bibinfo{person}{Dirk
  Weissenborn}, \bibinfo{person}{Xiaohua Zhai}, \bibinfo{person}{Thomas
  Unterthiner}, \bibinfo{person}{Mostafa Dehghani}, \bibinfo{person}{Matthias
  Minderer}, \bibinfo{person}{Georg Heigold}, \bibinfo{person}{Sylvain Gelly},
  {et~al\mbox{.}}} \bibinfo{year}{2020}\natexlab{}.
\newblock \showarticletitle{An image is worth 16x16 words: Transformers for
  image recognition at scale}.
\newblock \bibinfo{journal}{\emph{arXiv preprint arXiv:2010.11929}}
  (\bibinfo{year}{2020}).
\newblock


\bibitem[Feng et~al\mbox{.}(2025)]%
        {feng2025narrlv}
\bibfield{author}{\bibinfo{person}{Xiaokun Feng}, \bibinfo{person}{Haiming Yu},
  \bibinfo{person}{Meiqi Wu}, \bibinfo{person}{Shiyu Hu},
  \bibinfo{person}{Jintao Chen}, \bibinfo{person}{Chen Zhu},
  \bibinfo{person}{Jiahong Wu}, \bibinfo{person}{Xiangxiang Chu}, {and}
  \bibinfo{person}{Kaiqi Huang}.} \bibinfo{year}{2025}\natexlab{}.
\newblock \showarticletitle{NarrLV: Towards a Comprehensive Narrative-Centric
  Evaluation for Long Video Generation Models}.
\newblock \bibinfo{journal}{\emph{arXiv preprint arXiv:2507.11245}}
  (\bibinfo{year}{2025}).
\newblock


\bibitem[He et~al\mbox{.}(2020)]%
        {moco}
\bibfield{author}{\bibinfo{person}{Kaiming He}, \bibinfo{person}{Haoqi Fan},
  \bibinfo{person}{Yuxin Wu}, \bibinfo{person}{Saining Xie}, {and}
  \bibinfo{person}{Ross Girshick}.} \bibinfo{year}{2020}\natexlab{}.
\newblock \showarticletitle{Momentum contrast for unsupervised visual
  representation learning}. In \bibinfo{booktitle}{\emph{Proceedings of the
  IEEE/CVF conference on computer vision and pattern recognition}}.
  \bibinfo{pages}{9729--9738}.
\newblock


\bibitem[Jia et~al\mbox{.}(2021)]%
        {align}
\bibfield{author}{\bibinfo{person}{Chao Jia}, \bibinfo{person}{Yinfei Yang},
  \bibinfo{person}{Ye Xia}, \bibinfo{person}{Yi-Ting Chen},
  \bibinfo{person}{Zarana Parekh}, \bibinfo{person}{Hieu Pham},
  \bibinfo{person}{Quoc Le}, \bibinfo{person}{Yun-Hsuan Sung},
  \bibinfo{person}{Zhen Li}, {and} \bibinfo{person}{Tom Duerig}.}
  \bibinfo{year}{2021}\natexlab{}.
\newblock \showarticletitle{Scaling up visual and vision-language
  representation learning with noisy text supervision}. In
  \bibinfo{booktitle}{\emph{International conference on machine learning}}.
  PMLR, \bibinfo{pages}{4904--4916}.
\newblock


\bibitem[Kahneman et~al\mbox{.}(1993)]%
        {peak-end}
\bibfield{author}{\bibinfo{person}{Daniel Kahneman}, \bibinfo{person}{Barbara~L
  Fredrickson}, \bibinfo{person}{Charles~A Schreiber}, {and}
  \bibinfo{person}{Donald~A Redelmeier}.} \bibinfo{year}{1993}\natexlab{}.
\newblock \showarticletitle{When more pain is preferred to less: Adding a
  better end}.
\newblock \bibinfo{journal}{\emph{Psychological science}} \bibinfo{volume}{4},
  \bibinfo{number}{6} (\bibinfo{year}{1993}), \bibinfo{pages}{401--405}.
\newblock


\bibitem[Kuang et~al\mbox{.}(2019)]%
        {avaq6000}
\bibfield{author}{\bibinfo{person}{Qi Kuang}, \bibinfo{person}{Xin Jin},
  \bibinfo{person}{Qinping Zhao}, {and} \bibinfo{person}{Bin Zhou}.}
  \bibinfo{year}{2019}\natexlab{}.
\newblock \showarticletitle{Deep multimodality learning for UAV video aesthetic
  quality assessment}.
\newblock \bibinfo{journal}{\emph{IEEE Transactions on Multimedia}}
  \bibinfo{volume}{22}, \bibinfo{number}{10} (\bibinfo{year}{2019}),
  \bibinfo{pages}{2623--2634}.
\newblock


\bibitem[Kuehne et~al\mbox{.}(2011)]%
        {hmdb}
\bibfield{author}{\bibinfo{person}{Hildegard Kuehne}, \bibinfo{person}{Hueihan
  Jhuang}, \bibinfo{person}{Est{\'\i}baliz Garrote}, \bibinfo{person}{Tomaso
  Poggio}, {and} \bibinfo{person}{Thomas Serre}.}
  \bibinfo{year}{2011}\natexlab{}.
\newblock \showarticletitle{HMDB: a large video database for human motion
  recognition}. In \bibinfo{booktitle}{\emph{2011 International conference on
  computer vision}}. IEEE, \bibinfo{pages}{2556--2563}.
\newblock


\bibitem[Li et~al\mbox{.}(2022)]%
        {blip}
\bibfield{author}{\bibinfo{person}{Junnan Li}, \bibinfo{person}{Dongxu Li},
  \bibinfo{person}{Caiming Xiong}, {and} \bibinfo{person}{Steven Hoi}.}
  \bibinfo{year}{2022}\natexlab{}.
\newblock \showarticletitle{Blip: Bootstrapping language-image pre-training for
  unified vision-language understanding and generation}. In
  \bibinfo{booktitle}{\emph{International conference on machine learning}}.
  PMLR, \bibinfo{pages}{12888--12900}.
\newblock


\bibitem[Li et~al\mbox{.}(2024)]%
        {mvbench}
\bibfield{author}{\bibinfo{person}{Kunchang Li}, \bibinfo{person}{Yali Wang},
  \bibinfo{person}{Yinan He}, \bibinfo{person}{Yizhuo Li}, \bibinfo{person}{Yi
  Wang}, \bibinfo{person}{Yi Liu}, \bibinfo{person}{Zun Wang},
  \bibinfo{person}{Jilan Xu}, \bibinfo{person}{Guo Chen}, \bibinfo{person}{Ping
  Luo}, {et~al\mbox{.}}} \bibinfo{year}{2024}\natexlab{}.
\newblock \showarticletitle{Mvbench: A comprehensive multi-modal video
  understanding benchmark}. In \bibinfo{booktitle}{\emph{Proceedings of the
  IEEE/CVF Conference on Computer Vision and Pattern Recognition}}.
  \bibinfo{pages}{22195--22206}.
\newblock


\bibitem[Li et~al\mbox{.}(2025a)]%
        {li2025next}
\bibfield{author}{\bibinfo{person}{Mingxing Li}, \bibinfo{person}{Rui Wang},
  \bibinfo{person}{Lei Sun}, \bibinfo{person}{Yancheng Bai}, {and}
  \bibinfo{person}{Xiangxiang Chu}.} \bibinfo{year}{2025}\natexlab{a}.
\newblock \showarticletitle{Next Token Is Enough: Realistic Image Quality and
  Aesthetic Scoring with Multimodal Large Language Model}.
\newblock \bibinfo{journal}{\emph{arXiv preprint arXiv:2503.06141}}
  (\bibinfo{year}{2025}).
\newblock


\bibitem[Li et~al\mbox{.}(2025b)]%
        {qinsight}
\bibfield{author}{\bibinfo{person}{Weiqi Li}, \bibinfo{person}{Xuanyu Zhang},
  \bibinfo{person}{Shijie Zhao}, \bibinfo{person}{Yabin Zhang},
  \bibinfo{person}{Junlin Li}, \bibinfo{person}{Li Zhang}, {and}
  \bibinfo{person}{Jian Zhang}.} \bibinfo{year}{2025}\natexlab{b}.
\newblock \showarticletitle{Q-insight: Understanding image quality via visual
  reinforcement learning}.
\newblock \bibinfo{journal}{\emph{arXiv preprint arXiv:2503.22679}}
  (\bibinfo{year}{2025}).
\newblock


\bibitem[Ling et~al\mbox{.}(2025)]%
        {ling2025vmbench}
\bibfield{author}{\bibinfo{person}{Xinran Ling}, \bibinfo{person}{Chen Zhu},
  \bibinfo{person}{Meiqi Wu}, \bibinfo{person}{Hangyu Li},
  \bibinfo{person}{Xiaokun Feng}, \bibinfo{person}{Cundian Yang},
  \bibinfo{person}{Aiming Hao}, \bibinfo{person}{Jiashu Zhu},
  \bibinfo{person}{Jiahong Wu}, {and} \bibinfo{person}{Xiangxiang Chu}.}
  \bibinfo{year}{2025}\natexlab{}.
\newblock \showarticletitle{Vmbench: A benchmark for perception-aligned video
  motion generation}. In \bibinfo{booktitle}{\emph{Proceedings of the IEEE/CVF
  International Conference on Computer Vision}}. \bibinfo{pages}{13087--13098}.
\newblock


\bibitem[Loshchilov and Hutter(2017)]%
        {adamw}
\bibfield{author}{\bibinfo{person}{Ilya Loshchilov} {and}
  \bibinfo{person}{Frank Hutter}.} \bibinfo{year}{2017}\natexlab{}.
\newblock \showarticletitle{Decoupled weight decay regularization}.
\newblock \bibinfo{journal}{\emph{arXiv preprint arXiv:1711.05101}}
  (\bibinfo{year}{2017}).
\newblock


\bibitem[Luo et~al\mbox{.}(2022)]%
        {clip4clip}
\bibfield{author}{\bibinfo{person}{Huaishao Luo}, \bibinfo{person}{Lei Ji},
  \bibinfo{person}{Ming Zhong}, \bibinfo{person}{Yang Chen},
  \bibinfo{person}{Wen Lei}, \bibinfo{person}{Nan Duan}, {and}
  \bibinfo{person}{Tianrui Li}.} \bibinfo{year}{2022}\natexlab{}.
\newblock \showarticletitle{Clip4clip: An empirical study of clip for end to
  end video clip retrieval and captioning}.
\newblock \bibinfo{journal}{\emph{Neurocomputing}}  \bibinfo{volume}{508}
  (\bibinfo{year}{2022}), \bibinfo{pages}{293--304}.
\newblock


\bibitem[Luo and Tang(2008)]%
        {photovaa}
\bibfield{author}{\bibinfo{person}{Yiwen Luo} {and} \bibinfo{person}{Xiaoou
  Tang}.} \bibinfo{year}{2008}\natexlab{}.
\newblock \showarticletitle{Photo and video quality evaluation: Focusing on the
  subject}. In \bibinfo{booktitle}{\emph{European conference on computer
  vision}}. Springer, \bibinfo{pages}{386--399}.
\newblock


\bibitem[Ma et~al\mbox{.}(2022)]%
        {xclip}
\bibfield{author}{\bibinfo{person}{Yiwei Ma}, \bibinfo{person}{Guohai Xu},
  \bibinfo{person}{Xiaoshuai Sun}, \bibinfo{person}{Ming Yan},
  \bibinfo{person}{Ji Zhang}, {and} \bibinfo{person}{Rongrong Ji}.}
  \bibinfo{year}{2022}\natexlab{}.
\newblock \showarticletitle{X-clip: End-to-end multi-grained contrastive
  learning for video-text retrieval}. In \bibinfo{booktitle}{\emph{Proceedings
  of the 30th ACM international conference on multimedia}}.
  \bibinfo{pages}{638--647}.
\newblock


\bibitem[Maerten et~al\mbox{.}(2025)]%
        {lapis}
\bibfield{author}{\bibinfo{person}{Anne-Sofie Maerten}, \bibinfo{person}{Li-Wei
  Chen}, \bibinfo{person}{Stefanie De~Winter}, \bibinfo{person}{Christophe
  Bossens}, {and} \bibinfo{person}{Johan Wagemans}.}
  \bibinfo{year}{2025}\natexlab{}.
\newblock \showarticletitle{LAPIS: A novel dataset for personalized image
  aesthetic assessment}. In \bibinfo{booktitle}{\emph{Proceedings of the
  Computer Vision and Pattern Recognition Conference}}.
  \bibinfo{pages}{6302--6311}.
\newblock


\bibitem[Mi et~al\mbox{.}(2025)]%
        {qclip}
\bibfield{author}{\bibinfo{person}{Yachun Mi}, \bibinfo{person}{Yu Li},
  \bibinfo{person}{Yanting Li}, \bibinfo{person}{Chen Hui},
  \bibinfo{person}{Tong Zhang}, \bibinfo{person}{Zhixuan Li},
  \bibinfo{person}{Chenyue Song}, \bibinfo{person}{Wei Yang~Bryan Lim}, {and}
  \bibinfo{person}{Shaohui Liu}.} \bibinfo{year}{2025}\natexlab{}.
\newblock \showarticletitle{Q-CLIP: Unleashing the Power of Vision-Language
  Models for Video Quality Assessment through Unified Cross-Modal Adaptation}.
\newblock \bibinfo{journal}{\emph{arXiv preprint arXiv:2508.06092}}
  (\bibinfo{year}{2025}).
\newblock


\bibitem[Min et~al\mbox{.}(2024)]%
        {vqa}
\bibfield{author}{\bibinfo{person}{Xiongkuo Min}, \bibinfo{person}{Huiyu Duan},
  \bibinfo{person}{Wei Sun}, \bibinfo{person}{Yucheng Zhu}, {and}
  \bibinfo{person}{Guangtao Zhai}.} \bibinfo{year}{2024}\natexlab{}.
\newblock \showarticletitle{Perceptual video quality assessment: A survey}.
\newblock \bibinfo{journal}{\emph{Science China Information Sciences}}
  \bibinfo{volume}{67}, \bibinfo{number}{11} (\bibinfo{year}{2024}),
  \bibinfo{pages}{211301}.
\newblock


\bibitem[Moorthy et~al\mbox{.}(2010)]%
        {telefonica}
\bibfield{author}{\bibinfo{person}{Anush~K Moorthy}, \bibinfo{person}{Pere
  Obrador}, {and} \bibinfo{person}{Nuria Oliver}.}
  \bibinfo{year}{2010}\natexlab{}.
\newblock \showarticletitle{Towards computational models of the visual
  aesthetic appeal of consumer videos}. In \bibinfo{booktitle}{\emph{European
  conference on computer vision}}. Springer, \bibinfo{pages}{1--14}.
\newblock


\bibitem[Murray et~al\mbox{.}(2012)]%
        {ava}
\bibfield{author}{\bibinfo{person}{Naila Murray}, \bibinfo{person}{Luca
  Marchesotti}, {and} \bibinfo{person}{Florent Perronnin}.}
  \bibinfo{year}{2012}\natexlab{}.
\newblock \showarticletitle{AVA: A large-scale database for aesthetic visual
  analysis}. In \bibinfo{booktitle}{\emph{2012 IEEE conference on computer
  vision and pattern recognition}}. IEEE, \bibinfo{pages}{2408--2415}.
\newblock


\bibitem[Niu and Liu(2012)]%
        {professionalvideo}
\bibfield{author}{\bibinfo{person}{Yuzhen Niu} {and} \bibinfo{person}{Feng
  Liu}.} \bibinfo{year}{2012}\natexlab{}.
\newblock \showarticletitle{What makes a professional video? A computational
  aesthetics approach}.
\newblock \bibinfo{journal}{\emph{IEEE Transactions on Circuits and Systems for
  Video Technology}} \bibinfo{volume}{22}, \bibinfo{number}{7}
  (\bibinfo{year}{2012}), \bibinfo{pages}{1037--1049}.
\newblock


\bibitem[Qiao et~al\mbox{.}(2025)]%
        {vadb}
\bibfield{author}{\bibinfo{person}{Qianqian Qiao}, \bibinfo{person}{DanDan
  Zheng}, \bibinfo{person}{Yihang Bo}, \bibinfo{person}{Bao Peng},
  \bibinfo{person}{Heng Huang}, \bibinfo{person}{Longteng Jiang},
  \bibinfo{person}{Huaye Wang}, \bibinfo{person}{Jingdong Chen},
  \bibinfo{person}{Jun Zhou}, {and} \bibinfo{person}{Xin Jin}.}
  \bibinfo{year}{2025}\natexlab{}.
\newblock \showarticletitle{VADB: A Large-Scale Video Aesthetic Database with
  Professional and Multi-Dimensional Annotations}.
\newblock \bibinfo{journal}{\emph{arXiv preprint arXiv:2510.25238}}
  (\bibinfo{year}{2025}).
\newblock


\bibitem[Radford et~al\mbox{.}(2021)]%
        {clip}
\bibfield{author}{\bibinfo{person}{Alec Radford}, \bibinfo{person}{Jong~Wook
  Kim}, \bibinfo{person}{Chris Hallacy}, \bibinfo{person}{Aditya Ramesh},
  \bibinfo{person}{Gabriel Goh}, \bibinfo{person}{Sandhini Agarwal},
  \bibinfo{person}{Girish Sastry}, \bibinfo{person}{Amanda Askell},
  \bibinfo{person}{Pamela Mishkin}, \bibinfo{person}{Jack Clark},
  {et~al\mbox{.}}} \bibinfo{year}{2021}\natexlab{}.
\newblock \showarticletitle{Learning transferable visual models from natural
  language supervision}. In \bibinfo{booktitle}{\emph{International conference
  on machine learning}}. PMLR, \bibinfo{pages}{8748--8763}.
\newblock


\bibitem[Soomro et~al\mbox{.}(2012)]%
        {ucf101}
\bibfield{author}{\bibinfo{person}{Khurram Soomro},
  \bibinfo{person}{Amir~Roshan Zamir}, {and} \bibinfo{person}{Mubarak Shah}.}
  \bibinfo{year}{2012}\natexlab{}.
\newblock \showarticletitle{Ucf101: A dataset of 101 human actions classes from
  videos in the wild}.
\newblock \bibinfo{journal}{\emph{arXiv preprint arXiv:1212.0402}}
  (\bibinfo{year}{2012}).
\newblock


\bibitem[Sun et~al\mbox{.}(2022)]%
        {simplevqa}
\bibfield{author}{\bibinfo{person}{Wei Sun}, \bibinfo{person}{Xiongkuo Min},
  \bibinfo{person}{Wei Lu}, {and} \bibinfo{person}{Guangtao Zhai}.}
  \bibinfo{year}{2022}\natexlab{}.
\newblock \showarticletitle{A deep learning based no-reference quality
  assessment model for ugc videos}. In \bibinfo{booktitle}{\emph{Proceedings of
  the 30th ACM International Conference on Multimedia}}.
  \bibinfo{pages}{856--865}.
\newblock


\bibitem[Tang et~al\mbox{.}(2025)]%
        {videounderstand}
\bibfield{author}{\bibinfo{person}{Yunlong Tang}, \bibinfo{person}{Jing Bi},
  \bibinfo{person}{Siting Xu}, \bibinfo{person}{Luchuan Song},
  \bibinfo{person}{Susan Liang}, \bibinfo{person}{Teng Wang},
  \bibinfo{person}{Daoan Zhang}, \bibinfo{person}{Jie An},
  \bibinfo{person}{Jingyang Lin}, \bibinfo{person}{Rongyi Zhu},
  {et~al\mbox{.}}} \bibinfo{year}{2025}\natexlab{}.
\newblock \showarticletitle{Video understanding with large language models: A
  survey}.
\newblock \bibinfo{journal}{\emph{IEEE Transactions on Circuits and Systems for
  Video Technology}} (\bibinfo{year}{2025}).
\newblock


\bibitem[Tschannen et~al\mbox{.}(2025)]%
        {siglip2}
\bibfield{author}{\bibinfo{person}{Michael Tschannen}, \bibinfo{person}{Alexey
  Gritsenko}, \bibinfo{person}{Xiao Wang}, \bibinfo{person}{Muhammad~Ferjad
  Naeem}, \bibinfo{person}{Ibrahim Alabdulmohsin}, \bibinfo{person}{Nikhil
  Parthasarathy}, \bibinfo{person}{Talfan Evans}, \bibinfo{person}{Lucas
  Beyer}, \bibinfo{person}{Ye Xia}, \bibinfo{person}{Basil Mustafa},
  {et~al\mbox{.}}} \bibinfo{year}{2025}\natexlab{}.
\newblock \showarticletitle{Siglip 2: Multilingual vision-language encoders
  with improved semantic understanding, localization, and dense features}.
\newblock \bibinfo{journal}{\emph{arXiv preprint arXiv:2502.14786}}
  (\bibinfo{year}{2025}).
\newblock


\bibitem[Tzelepis et~al\mbox{.}(2016)]%
        {vqa700}
\bibfield{author}{\bibinfo{person}{Christos Tzelepis},
  \bibinfo{person}{Eftichia Mavridaki}, \bibinfo{person}{Vasileios Mezaris},
  {and} \bibinfo{person}{Ioannis Patras}.} \bibinfo{year}{2016}\natexlab{}.
\newblock \showarticletitle{Video aesthetic quality assessment using kernel
  Support Vector Machine with isotropic Gaussian sample uncertainty
  (KSVM-IGSU)}. In \bibinfo{booktitle}{\emph{2016 IEEE International Conference
  on Image Processing (ICIP)}}. IEEE, \bibinfo{pages}{2410--2414}.
\newblock


\bibitem[Vaswani et~al\mbox{.}(2017)]%
        {transformer}
\bibfield{author}{\bibinfo{person}{Ashish Vaswani}, \bibinfo{person}{Noam
  Shazeer}, \bibinfo{person}{Niki Parmar}, \bibinfo{person}{Jakob Uszkoreit},
  \bibinfo{person}{Llion Jones}, \bibinfo{person}{Aidan~N Gomez},
  \bibinfo{person}{{\L}ukasz Kaiser}, {and} \bibinfo{person}{Illia
  Polosukhin}.} \bibinfo{year}{2017}\natexlab{}.
\newblock \showarticletitle{Attention is all you need}.
\newblock \bibinfo{journal}{\emph{Advances in neural information processing
  systems}}  \bibinfo{volume}{30} (\bibinfo{year}{2017}).
\newblock


\bibitem[Wang et~al\mbox{.}(2021)]%
        {actionclip}
\bibfield{author}{\bibinfo{person}{Mengmeng Wang}, \bibinfo{person}{Jiazheng
  Xing}, {and} \bibinfo{person}{Yong Liu}.} \bibinfo{year}{2021}\natexlab{}.
\newblock \showarticletitle{Actionclip: A new paradigm for video action
  recognition}.
\newblock \bibinfo{journal}{\emph{arXiv preprint arXiv:2109.08472}}
  (\bibinfo{year}{2021}).
\newblock


\bibitem[Wen et~al\mbox{.}(2024)]%
        {modularbvqa}
\bibfield{author}{\bibinfo{person}{Wen Wen}, \bibinfo{person}{Mu Li},
  \bibinfo{person}{Yabin Zhang}, \bibinfo{person}{Yiting Liao},
  \bibinfo{person}{Junlin Li}, \bibinfo{person}{Li Zhang}, {and}
  \bibinfo{person}{Kede Ma}.} \bibinfo{year}{2024}\natexlab{}.
\newblock \showarticletitle{Modular blind video quality assessment}. In
  \bibinfo{booktitle}{\emph{Proceedings of the IEEE/CVF Conference on Computer
  Vision and Pattern Recognition}}. \bibinfo{pages}{2763--2772}.
\newblock


\bibitem[Wu et~al\mbox{.}(2022)]%
        {fastvqa}
\bibfield{author}{\bibinfo{person}{Haoning Wu}, \bibinfo{person}{Chaofeng
  Chen}, \bibinfo{person}{Jingwen Hou}, \bibinfo{person}{Liang Liao},
  \bibinfo{person}{Annan Wang}, \bibinfo{person}{Wenxiu Sun},
  \bibinfo{person}{Qiong Yan}, {and} \bibinfo{person}{Weisi Lin}.}
  \bibinfo{year}{2022}\natexlab{}.
\newblock \showarticletitle{Fast-vqa: Efficient end-to-end video quality
  assessment with fragment sampling}. In \bibinfo{booktitle}{\emph{European
  conference on computer vision}}. Springer, \bibinfo{pages}{538--554}.
\newblock


\bibitem[Wu et~al\mbox{.}(2023b)]%
        {explain}
\bibfield{author}{\bibinfo{person}{H. Wu}, \bibinfo{person}{E. Zhang},
  \bibinfo{person}{L. Liao}, {et~al\mbox{.}}} \bibinfo{year}{2023}\natexlab{b}.
\newblock \showarticletitle{Towards explainable in-the-wild video quality
  assessment: a database and a language-prompted approach}.
\newblock \bibinfo{journal}{\emph{Proceedings of the 31st acm international
  conference on multimedia}} (\bibinfo{year}{2023}),
  \bibinfo{pages}{1045--1054}.
\newblock


\bibitem[Wu et~al\mbox{.}(2023a)]%
        {divide}
\bibfield{author}{\bibinfo{person}{Haoning Wu}, \bibinfo{person}{Erli Zhang},
  \bibinfo{person}{Liang Liao}, \bibinfo{person}{Chaofeng Chen},
  \bibinfo{person}{Jingwen Hou}, \bibinfo{person}{Annan Wang},
  \bibinfo{person}{Wenxiu Sun}, \bibinfo{person}{Qiong Yan}, {and}
  \bibinfo{person}{Weisi Lin}.} \bibinfo{year}{2023}\natexlab{a}.
\newblock \showarticletitle{Exploring video quality assessment on user
  generated contents from aesthetic and technical perspectives}. In
  \bibinfo{booktitle}{\emph{Proceedings of the IEEE/CVF international
  conference on computer vision}}. \bibinfo{pages}{20144--20154}.
\newblock


\bibitem[Wu et~al\mbox{.}(2023c)]%
        {qalign}
\bibfield{author}{\bibinfo{person}{Haoning Wu}, \bibinfo{person}{Zicheng
  Zhang}, \bibinfo{person}{Weixia Zhang}, \bibinfo{person}{Chaofeng Chen},
  \bibinfo{person}{Liang Liao}, \bibinfo{person}{Chunyi Li},
  \bibinfo{person}{Yixuan Gao}, \bibinfo{person}{Annan Wang},
  \bibinfo{person}{Erli Zhang}, \bibinfo{person}{Wenxiu Sun}, {et~al\mbox{.}}}
  \bibinfo{year}{2023}\natexlab{c}.
\newblock \showarticletitle{Q-align: Teaching lmms for visual scoring via
  discrete text-defined levels}.
\newblock \bibinfo{journal}{\emph{arXiv preprint arXiv:2312.17090}}
  (\bibinfo{year}{2023}).
\newblock


\bibitem[Wu et~al\mbox{.}(2025)]%
        {vqr1}
\bibfield{author}{\bibinfo{person}{Tianhe Wu}, \bibinfo{person}{Jian Zou},
  \bibinfo{person}{Jie Liang}, \bibinfo{person}{Lei Zhang}, {and}
  \bibinfo{person}{Kede Ma}.} \bibinfo{year}{2025}\natexlab{}.
\newblock \showarticletitle{Visualquality-r1: Reasoning-induced image quality
  assessment via reinforcement learning to rank}.
\newblock \bibinfo{journal}{\emph{arXiv preprint arXiv:2505.14460}}
  (\bibinfo{year}{2025}).
\newblock


\bibitem[Xie et~al\mbox{.}(2026)]%
        {xie2026qhawkeye}
\bibfield{author}{\bibinfo{person}{Wulin Xie}, \bibinfo{person}{Rui Dai},
  \bibinfo{person}{Ruidong Ding}, \bibinfo{person}{Kaikui Liu},
  \bibinfo{person}{Xiangxiang Chu}, \bibinfo{person}{Xinwen Hou}, {and}
  \bibinfo{person}{Jie Wen}.} \bibinfo{year}{2026}\natexlab{}.
\newblock \showarticletitle{Q-Hawkeye: Reliable Visual Policy Optimization for
  Image Quality Assessment}.
\newblock \bibinfo{journal}{\emph{arXiv preprint arXiv:2601.22920}}
  (\bibinfo{year}{2026}).
\newblock


\bibitem[Xing et~al\mbox{.}(2024)]%
        {clipvqa}
\bibfield{author}{\bibinfo{person}{Fengchuang Xing}, \bibinfo{person}{Mingjie
  Li}, \bibinfo{person}{Yuan-Gen Wang}, \bibinfo{person}{Guopu Zhu}, {and}
  \bibinfo{person}{Xiaochun Cao}.} \bibinfo{year}{2024}\natexlab{}.
\newblock \showarticletitle{Clipvqa: Video quality assessment via clip}.
\newblock \bibinfo{journal}{\emph{IEEE Transactions on Broadcasting}}
  \bibinfo{volume}{71}, \bibinfo{number}{1} (\bibinfo{year}{2024}),
  \bibinfo{pages}{291--306}.
\newblock


\bibitem[Xu et~al\mbox{.}(2021)]%
        {videoclip}
\bibfield{author}{\bibinfo{person}{Hu Xu}, \bibinfo{person}{Gargi Ghosh},
  \bibinfo{person}{Po-Yao Huang}, \bibinfo{person}{Dmytro Okhonko},
  \bibinfo{person}{Armen Aghajanyan}, \bibinfo{person}{Florian Metze},
  \bibinfo{person}{Luke Zettlemoyer}, {and} \bibinfo{person}{Christoph
  Feichtenhofer}.} \bibinfo{year}{2021}\natexlab{}.
\newblock \showarticletitle{Videoclip: Contrastive pre-training for zero-shot
  video-text understanding}. In \bibinfo{booktitle}{\emph{Proceedings of the
  2021 conference on empirical methods in natural language processing}}.
  \bibinfo{pages}{6787--6800}.
\newblock


\bibitem[Yuan et~al\mbox{.}(2024)]%
        {ptmvqa}
\bibfield{author}{\bibinfo{person}{Kun Yuan}, \bibinfo{person}{Hongbo Liu},
  \bibinfo{person}{Mading Li}, \bibinfo{person}{Muyi Sun},
  \bibinfo{person}{Ming Sun}, \bibinfo{person}{Jiachao Gong},
  \bibinfo{person}{Jinhua Hao}, \bibinfo{person}{Chao Zhou}, {and}
  \bibinfo{person}{Yansong Tang}.} \bibinfo{year}{2024}\natexlab{}.
\newblock \showarticletitle{Ptm-vqa: efficient video quality assessment
  leveraging diverse pretrained models from the wild}. In
  \bibinfo{booktitle}{\emph{Proceedings of the IEEE/CVF Conference on Computer
  Vision and Pattern Recognition}}. \bibinfo{pages}{2835--2845}.
\newblock


\bibitem[Zhai et~al\mbox{.}(2023)]%
        {siglip}
\bibfield{author}{\bibinfo{person}{Xiaohua Zhai}, \bibinfo{person}{Basil
  Mustafa}, \bibinfo{person}{Alexander Kolesnikov}, {and}
  \bibinfo{person}{Lucas Beyer}.} \bibinfo{year}{2023}\natexlab{}.
\newblock \showarticletitle{Sigmoid loss for language image pre-training}. In
  \bibinfo{booktitle}{\emph{Proceedings of the IEEE/CVF international
  conference on computer vision}}. \bibinfo{pages}{11975--11986}.
\newblock


\end{thebibliography}



\begin{thebibliography}{9}


\ifx \showCODEN    \undefined \def \showCODEN     #1{\unskip}     \fi
\ifx \showISBNx    \undefined \def \showISBNx     #1{\unskip}     \fi
\ifx \showISBNxiii \undefined \def \showISBNxiii  #1{\unskip}     \fi
\ifx \showISSN     \undefined \def \showISSN      #1{\unskip}     \fi
\ifx \showLCCN     \undefined \def \showLCCN      #1{\unskip}     \fi
\ifx \shownote     \undefined \def \shownote      #1{#1}          \fi
\ifx \showarticletitle \undefined \def \showarticletitle #1{#1}   \fi
\ifx \showURL      \undefined \def \showURL       {\relax}        \fi
\providecommand\bibfield[2]{#2}
\providecommand\bibinfo[2]{#2}
\providecommand\natexlab[1]{#1}
\providecommand\showeprint[2][]{arXiv:#2}

\bibitem[Dosovitskiy et~al\mbox{.}(2020)]%
        {vit}
\bibfield{author}{\bibinfo{person}{Alexey Dosovitskiy}, \bibinfo{person}{Lucas
  Beyer}, \bibinfo{person}{Alexander Kolesnikov}, \bibinfo{person}{Dirk
  Weissenborn}, \bibinfo{person}{Xiaohua Zhai}, \bibinfo{person}{Thomas
  Unterthiner}, \bibinfo{person}{Mostafa Dehghani}, \bibinfo{person}{Matthias
  Minderer}, \bibinfo{person}{Georg Heigold}, \bibinfo{person}{Sylvain Gelly},
  {et~al\mbox{.}}} \bibinfo{year}{2020}\natexlab{}.
\newblock \showarticletitle{An image is worth 16x16 words: Transformers for
  image recognition at scale}.
\newblock \bibinfo{journal}{\emph{arXiv preprint arXiv:2010.11929}}
  (\bibinfo{year}{2020}).
\newblock


\bibitem[Kay et~al\mbox{.}(2017)]%
        {k400}
\bibfield{author}{\bibinfo{person}{Will Kay}, \bibinfo{person}{Joao Carreira},
  \bibinfo{person}{Karen Simonyan}, \bibinfo{person}{Brian Zhang},
  \bibinfo{person}{Chloe Hillier}, \bibinfo{person}{Sudheendra
  Vijayanarasimhan}, \bibinfo{person}{Fabio Viola}, \bibinfo{person}{Tim
  Green}, \bibinfo{person}{Trevor Back}, \bibinfo{person}{Paul Natsev},
  {et~al\mbox{.}}} \bibinfo{year}{2017}\natexlab{}.
\newblock \showarticletitle{The kinetics human action video dataset}.
\newblock \bibinfo{journal}{\emph{arXiv preprint arXiv:1705.06950}}
  (\bibinfo{year}{2017}).
\newblock


\bibitem[Murray et~al\mbox{.}(2012)]%
        {ava}
\bibfield{author}{\bibinfo{person}{Naila Murray}, \bibinfo{person}{Luca
  Marchesotti}, {and} \bibinfo{person}{Florent Perronnin}.}
  \bibinfo{year}{2012}\natexlab{}.
\newblock \showarticletitle{AVA: A large-scale database for aesthetic visual
  analysis}. In \bibinfo{booktitle}{\emph{2012 IEEE conference on computer
  vision and pattern recognition}}. IEEE, \bibinfo{pages}{2408--2415}.
\newblock


\bibitem[Qiao et~al\mbox{.}(2025)]%
        {vadb}
\bibfield{author}{\bibinfo{person}{Qianqian Qiao}, \bibinfo{person}{DanDan
  Zheng}, \bibinfo{person}{Yihang Bo}, \bibinfo{person}{Bao Peng},
  \bibinfo{person}{Heng Huang}, \bibinfo{person}{Longteng Jiang},
  \bibinfo{person}{Huaye Wang}, \bibinfo{person}{Jingdong Chen},
  \bibinfo{person}{Jun Zhou}, {and} \bibinfo{person}{Xin Jin}.}
  \bibinfo{year}{2025}\natexlab{}.
\newblock \showarticletitle{VADB: A Large-Scale Video Aesthetic Database with
  Professional and Multi-Dimensional Annotations}.
\newblock \bibinfo{journal}{\emph{arXiv preprint arXiv:2510.25238}}
  (\bibinfo{year}{2025}).
\newblock


\bibitem[Radford et~al\mbox{.}(2021)]%
        {clip}
\bibfield{author}{\bibinfo{person}{Alec Radford}, \bibinfo{person}{Jong~Wook
  Kim}, \bibinfo{person}{Chris Hallacy}, \bibinfo{person}{Aditya Ramesh},
  \bibinfo{person}{Gabriel Goh}, \bibinfo{person}{Sandhini Agarwal},
  \bibinfo{person}{Girish Sastry}, \bibinfo{person}{Amanda Askell},
  \bibinfo{person}{Pamela Mishkin}, \bibinfo{person}{Jack Clark},
  {et~al\mbox{.}}} \bibinfo{year}{2021}\natexlab{}.
\newblock \showarticletitle{Learning transferable visual models from natural
  language supervision}. In \bibinfo{booktitle}{\emph{International conference
  on machine learning}}. PMLR, \bibinfo{pages}{8748--8763}.
\newblock


\bibitem[Thomee et~al\mbox{.}(2016)]%
        {yfcc}
\bibfield{author}{\bibinfo{person}{Bart Thomee}, \bibinfo{person}{David~A
  Shamma}, \bibinfo{person}{Gerald Friedland}, \bibinfo{person}{Benjamin
  Elizalde}, \bibinfo{person}{Karl Ni}, \bibinfo{person}{Douglas Poland},
  \bibinfo{person}{Damian Borth}, {and} \bibinfo{person}{Li-Jia Li}.}
  \bibinfo{year}{2016}\natexlab{}.
\newblock \showarticletitle{Yfcc100m: The new data in multimedia research}.
\newblock \bibinfo{journal}{\emph{Commun. ACM}} \bibinfo{volume}{59},
  \bibinfo{number}{2} (\bibinfo{year}{2016}), \bibinfo{pages}{64--73}.
\newblock


\bibitem[Wu et~al\mbox{.}(2023a)]%
        {divide}
\bibfield{author}{\bibinfo{person}{Haoning Wu}, \bibinfo{person}{Erli Zhang},
  \bibinfo{person}{Liang Liao}, \bibinfo{person}{Chaofeng Chen},
  \bibinfo{person}{Jingwen Hou}, \bibinfo{person}{Annan Wang},
  \bibinfo{person}{Wenxiu Sun}, \bibinfo{person}{Qiong Yan}, {and}
  \bibinfo{person}{Weisi Lin}.} \bibinfo{year}{2023}\natexlab{a}.
\newblock \showarticletitle{Exploring video quality assessment on user
  generated contents from aesthetic and technical perspectives}. In
  \bibinfo{booktitle}{\emph{Proceedings of the IEEE/CVF international
  conference on computer vision}}. \bibinfo{pages}{20144--20154}.
\newblock


\bibitem[Wu et~al\mbox{.}(2023b)]%
        {qalign}
\bibfield{author}{\bibinfo{person}{Haoning Wu}, \bibinfo{person}{Zicheng
  Zhang}, \bibinfo{person}{Weixia Zhang}, \bibinfo{person}{Chaofeng Chen},
  \bibinfo{person}{Liang Liao}, \bibinfo{person}{Chunyi Li},
  \bibinfo{person}{Yixuan Gao}, \bibinfo{person}{Annan Wang},
  \bibinfo{person}{Erli Zhang}, \bibinfo{person}{Wenxiu Sun}, {et~al\mbox{.}}}
  \bibinfo{year}{2023}\natexlab{b}.
\newblock \showarticletitle{Q-align: Teaching lmms for visual scoring via
  discrete text-defined levels}.
\newblock \bibinfo{journal}{\emph{arXiv preprint arXiv:2312.17090}}
  (\bibinfo{year}{2023}).
\newblock


\bibitem[Ye et~al\mbox{.}(2024)]%
        {mplug-owl2}
\bibfield{author}{\bibinfo{person}{Qinghao Ye}, \bibinfo{person}{Haiyang Xu},
  \bibinfo{person}{Jiabo Ye}, \bibinfo{person}{Ming Yan},
  \bibinfo{person}{Anwen Hu}, \bibinfo{person}{Haowei Liu}, \bibinfo{person}{Qi
  Qian}, \bibinfo{person}{Ji Zhang}, {and} \bibinfo{person}{Fei Huang}.}
  \bibinfo{year}{2024}\natexlab{}.
\newblock \showarticletitle{mplug-owl2: Revolutionizing multi-modal large
  language model with modality collaboration}. In
  \bibinfo{booktitle}{\emph{Proceedings of the ieee/cvf conference on computer
  vision and pattern recognition}}. \bibinfo{pages}{13040--13051}.
\newblock


\end{thebibliography}

\end{document}


\title{Supplementary Material\\ Peak-End-Net: A Peak-End Rule Inspired Framework for Generalizable Video Aesthetic Assessment}

\author{Geng Li}
\authornote{Both authors contributed equally to this work.}
\email{xiaofeng.lg@alibaba-inc.com}
\affiliation{
  \institution{AMAP, Alibaba Group}
  \city{Beijing}
  \country{China}
}

\author{Haiwen Li}
\authornotemark[1]
\email{lihaiwen.lhw@alibaba-inc.com}
\affiliation{
  \institution{AMAP, Alibaba Group}
  \city{Beijing}
  \country{China}
}

\author{Rui Chen}
\email{chenrui.chen@alibaba-inc.com}
\affiliation{
  \institution{AMAP, Alibaba Group}
  \city{Beijing}
  \country{China}
}

\author{Jing Tang}
\email{guangyu.tj@alibaba-inc.com}
\affiliation{
  \institution{AMAP, Alibaba Group}
  \city{Beijing}
  \country{China}
}

\author{Lei Sun}
\email{ally.sl@alibaba-inc.com}
\affiliation{
  \institution{AMAP, Alibaba Group}
  \city{Beijing}
  \country{China}
}

\author{Xiangxiang Chu}
\email{chuxiangxiang.cxx@alibaba-inc.com}
\affiliation{
  \institution{AMAP, Alibaba Group}
  \city{Beijing}
  \country{China}
}

\renewcommand{\shortauthors}{Geng Li et al.}

\maketitle

\appendix
This appendix mainly includes a detailed introduction to the benchmarks involved in the main paper (Sec.~\ref{A}), the specific implementations of the proposed modules (Sec.~\ref{B}), computational efficiency analysis (Sec.~~\ref{C}), and additional visualizations (Sec.~~\ref{D}).
\section{Benchmark Details}
\label{A}
\textbf{Video Aesthetic Database (VADB)}~\cite{vadb} is a large-scale dataset for video aesthetic assessment (VAA). It was proposed to address the lack of standardized and richly annotated benchmarks in this area. The dataset contains 10,490 videos collected from diverse sources, including films, TV dramas, documentaries, news, user-generated content, and AI-generated videos. A small portion of the videos has not yet been publicly released due to copyright restrictions. Each clip is 5–20 seconds long and belongs to one of four categories: character, natural scenery, architecture, and food.

\begin{figure}
\centering
\includegraphics[width=\columnwidth]{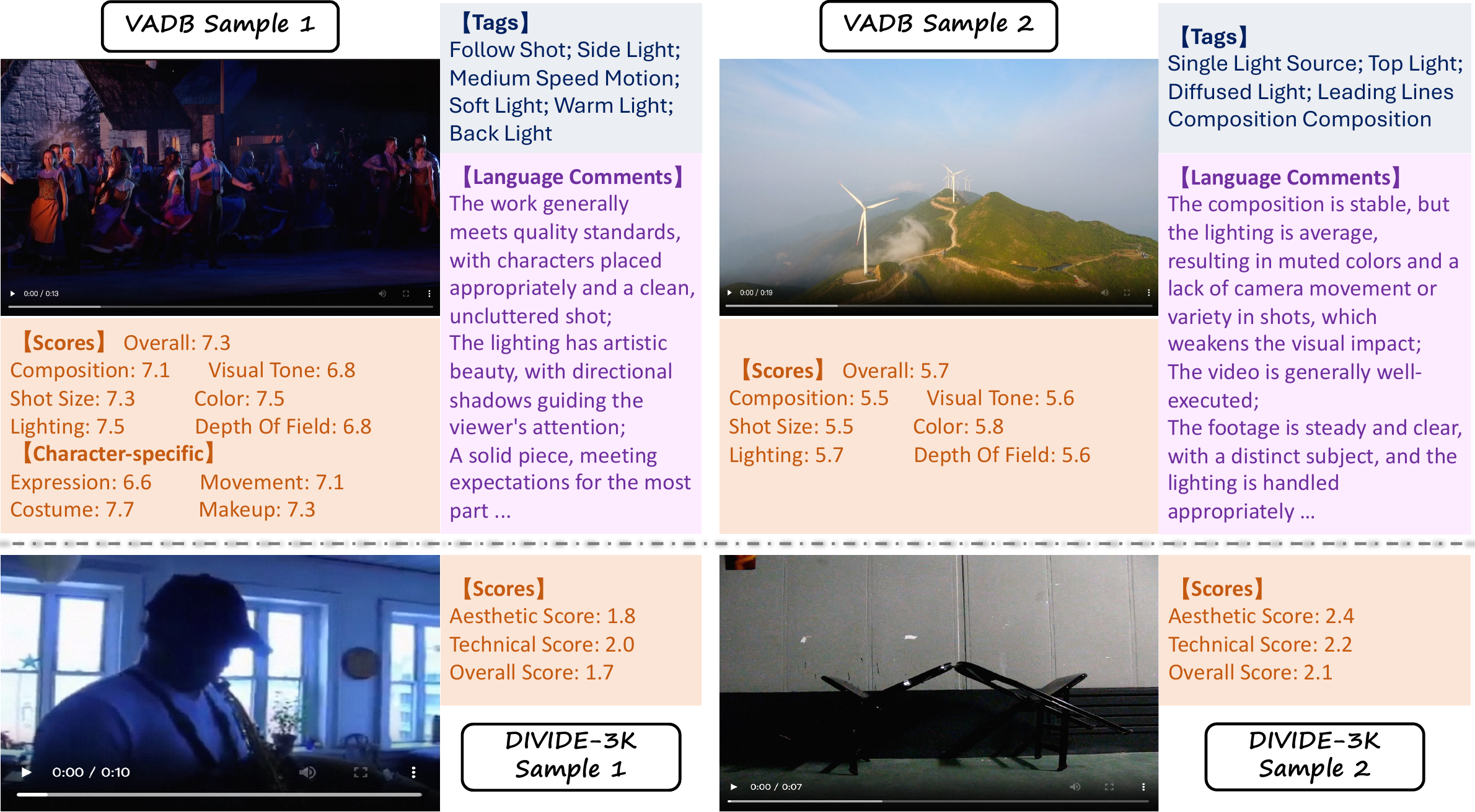}
\caption{Some representative samples: the two above are from VADB~\cite{vadb}, and in our experiments we only used the scores; the two below are from DIVIDE-3K~\cite{divide}, and we only evaluated the aesthetic dimension, i.e., the Aesthetic Score.}
\Description{vadb and divide}
\label{fig:vade_divide}
\end{figure}

A major feature of VADB is its professional and multi-dimensional annotation scheme. The dataset was annotated by 37 trained professionals with backgrounds in film, television, and media studies, under a carefully designed aesthetic framework developed by experts from the Beijing Film Academy. Each video has annotations from at least 13 annotators, including aesthetic scores, language comments, and technical tags. VADB provides an overall aesthetic score as well as attribute-level scores. The annotated attributes include composition, shot size, lighting, visual tone, color, depth of field, expression, movement, costume, and makeup. It also includes objective tags related to camera movement, composition, and lighting. Representative samples are shown in Figure~\ref{fig:vade_divide}(Top).

Compared with previous video aesthetic datasets, VADB is larger and more comprehensive in both scale and annotation richness. By combining numerical scores, textual comments, and technical tags, it offers a valuable benchmark for research on fine-grained video aesthetic assessment.

\textbf{Disentangled Video Quality Database (DIVIDE-3K)}~\cite{divide} is a large-scale user-generated content video quality assessment (UGC-VQA) database proposed to study video quality from both technical and aesthetic perspectives. Unlike previous UGC-VQA datasets that usually provide only one overall quality score, DIVIDE-3K explicitly disentangles how people judge videos in terms of visual distortions and content preference. The dataset contains 3,590 real-world videos collected mainly from YFCC-100M~\cite{yfcc} and Kinetics-400~\cite{k400}, covering diverse scenes, content, and quality levels. To ensure reliable annotations, the authors conducted an in-lab subjective study with 35 trained annotators, rather than relying on crowdsourcing. In total, the dataset includes about 450,000 human opinions, including overall quality scores, aesthetic scores, technical scores, and additional subjective reasoning labels that indicate how much each perspective influences the final overall judgment.

A key finding enabled by DIVIDE-3K is that human perception of UGC video quality is almost always shaped by both perspectives: a technically clean video may still be rated poorly if its content is unappealing, while an aesthetically attractive video may receive favorable scores despite distortions. Therefore, DIVIDE-3K provides a more comprehensive benchmark for understanding human quality perception and for developing explainable VQA models, such as DOVER, that can predict both overall quality and perspective-specific quality. Two samples are shown in Figure~\ref{fig:vade_divide}.

\section{Detailed Architectures}
\label{B}
The visual backbone we use is ViT-L/14~\cite{vit}, initialized with CLIP~\cite{clip} pretrained weights, whose output frame-level visual features are denoted as \(\mathbf{F}_v=\{\mathbf{f}_t \in \mathbb{R}^d\}_{t=1}^T\), where \(T=12\) by default and \(d=768\).

\subsection{IAA Head \texorpdfstring{$\Phi_{\text{IAA}}$}{}}
The IAA head \(\Phi_{\text{IAA}}\) is pretrained on the AVA~\cite{ava} dataset and then kept frozen and applied to the frame-level visual features. Given $\mathbf{f}_t\in\mathbb{R}^{768}$, the head predicts a 10-dimensional logit vector:
\begin{equation}
\mathbf{z}_t=\Phi_{\text{IAA}}(\mathbf{f}_t)\in\mathbb{R}^{10}.
\end{equation}
$\Phi_{\text{IAA}}$ is implemented as an MLP with dimensions \texttt{768} $\rightarrow$ \texttt{512} $\rightarrow$ \texttt{256} $\rightarrow$ \texttt{10}, where the first hidden layer is followed by \texttt{LayerNorm}, \texttt{ReLU}, and \texttt{Dropout}, and the second hidden layer is followed by \texttt{ReLU} and \texttt{Dropout}. The dropout ratio is $0.3$. The logits are normalized by softmax to obtain a 10-way aesthetic distribution, which is used to compute a weighted aesthetic score $s_t$ for each sampled frame. 

\subsection{Rhythm Encoder \texorpdfstring{$\Phi_{\text{RE}}$}{}}
The rhythm encoder $\Phi_{\text{RE}}$ models temporal variation patterns in the frame-wise aesthetic score sequence $\mathbf{s}=[s_1,\dots,s_T]\in\mathbb R^{1\times T}$. 
It treats $\mathbf{s}$ as a 1D temporal signal and applies three parallel 1D convolution branches with kernel sizes 3, 5, and 7, respectively. Each branch has channel dimensions $\texttt{1} \rightarrow \texttt{16} \rightarrow \texttt{32}$ with \texttt{GELU} activations. The three outputs are concatenated into a $96\times T$ feature map and fused by a $1\times1$ convolution into a $64\times T$ representation. 
After \texttt{Mean-pooling} over time, a \texttt{linear Projection} and \texttt{LayerNorm} are applied to obtain the final rhythm feature ($d'=64$):
\begin{equation}
\mathbf{f}_{\text{rhythm}}=\Phi_{\text{RE}}(\mathbf{s})\in\mathbb{R}^{d'}.
\end{equation}

\begin{figure*}
\centering
\includegraphics[width=\textwidth]{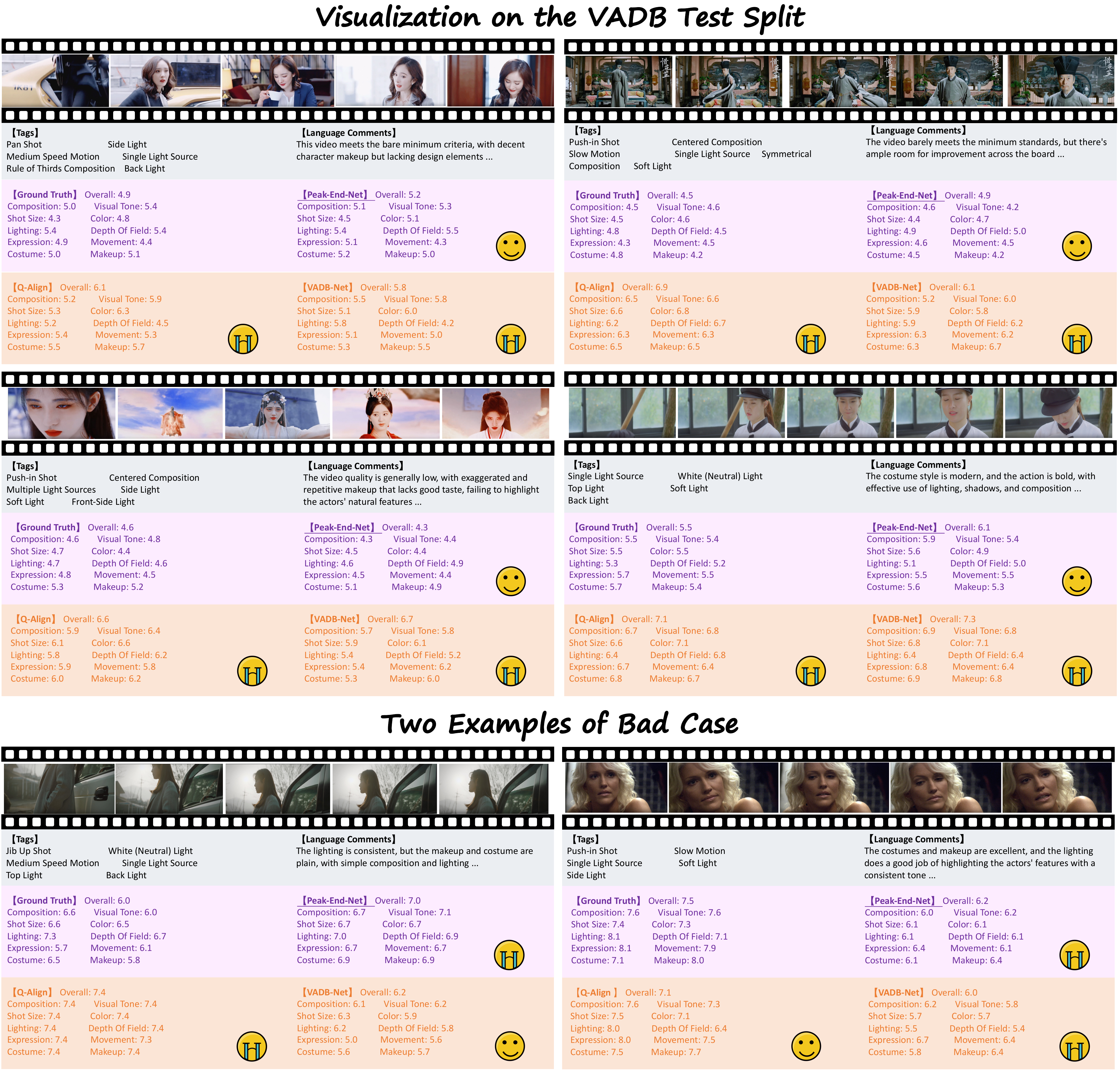}
\caption{Qualitative results on VADB. We present the tags, language comments, ground-truth annotations, and the predicted overall and attribute-level scores produced by \textit{Peak-End-Net}, \textit{Q-Align}, and \textit{VADB-Net}. We additionally provide an analysis of failure cases.}
\Description{case analysis}
\label{fig:case}
\end{figure*}

\subsection{VAA Head \texorpdfstring{$\Phi_{\text{VAA}}$}{}}
The VAA head $\Phi_{\text{VAA}}$ predicts the overall and attribute-level scores from the concatenated representation:
\[
\mathbf{f}_{\text{concat}}=\mathbf{f}_{\text{video}}\oplus\mathbf{f}_{\text{rhythm}}.
\]
It consists of a shared MLP backbone with dimensions $(\texttt{768+64}) \rightarrow \texttt{512} \rightarrow \texttt{256}$, followed by \texttt{LayerNorm}, where \texttt{GELU} activations are used after each layer and \texttt{dropout} with ratio $0.1$ is applied after the first layer. Based on the shared feature, we use $11$ attribute-specific regression heads, each implemented as $\texttt{256} \rightarrow \texttt{128} \rightarrow \texttt{1}$. The outputs of all heads are concatenated into an 11-dimensional vector, followed by a sigmoid function to obtain normalized scores, which are used to compute $\mathcal L_\text{VAA}$.

\subsection{Gate Fusion Module \texorpdfstring{$\Phi_{\text{GF}}$}{}}
The gate fusion module $\Phi_{\text{GF}}$ predicts a sample-wise scalar gate $\lambda$ for combining the predicted VAA overall score and the global IAA score. It takes $\mathbf{f}_{\text{video}}\oplus\mathbf{f}_{\text{rhythm}}$ as input and is implemented as a two-layer MLP with dimensions $(\texttt{768+64}) \rightarrow \texttt{128} \rightarrow \texttt{1}$, followed by \texttt{GELU}, \texttt{Dropout} ($p=0.1$), and a final sigmoid activation. The fused overall score is computed as:
\begin{equation}
\hat{y}^{\,\text{fusion}}_{\text{overall}}
=
\lambda\,\hat{y}_{\text{overall}}
+
(1-\lambda)\,\hat{y}_{\text{IAA}},
\end{equation}
where $\lambda\in[0,1]$ is the predicted gate value. 

\subsection{Temporal Transformer Baseline}
For the temporal transformer baseline in Table 4 of the main paper, we replace the proposed \textit{peak-end} aggregation with a standard transformer-based temporal encoder over frame-level features. A learnable \text{[CLS]} token is prepended to the frame sequence, and learnable temporal positional embeddings are added before the tokens are fed into the encoder. The model uses $4$ transformer encoder layers, $8$ attention heads, hidden dimension $768$, and a  feed-forward dimension $2048$, with \texttt{LayerNorm} and \texttt{Dropout} ($p=0.1$). The video representation is taken as the output of the \text{[CLS]} token.

\section{Computational Efficiency}
\label{C}
Another advantage of our framework is its parameter-efficient training scheme. Compared with VADB-Net~\cite{vadb}, which involves training the visual and textual backbones, and Q-Align~\cite{qalign}, which adapts a much larger multimodal large language model~\cite{mplug-owl2}, our method keeps the ViT backbone frozen and uses a frozen AVA-pretrained IAA head. Only lightweight modules, including $\Phi_{\text{VAA}}$, $\Phi_{\text{RE}}$, and $\Phi_{\text{GF}}$ are optimized during training.

Specifically, the total number of trainable parameters in our method is only \textbf{1.05M}, accounting for \textbf{less than 1\%} of the total parameter count. The number of trainable parameters is much smaller than that of VADB-Net and Q-Align, leading to lower training cost and memory demand while achieving stronger performance.

\section{Additional Qualitative Results}
\label{D}
Figure~\ref{fig:case} presents additional qualitative examples on VADB. \ding{182} \textit{Advantage of Peak-End-Net.} In most cases, \textit{Peak-End-Net} produces predictions that are noticeably closer to the ground truth than \textit{Q-Align}~\cite{qalign} and \textit{VADB-Net}~\cite{vadb}, especially for videos with large frame-wise variation. When the visual quality changes significantly across time, our method can better capture the contribution of salient moments and the ending segment, leading to more accurate overall and attribute-level assessment. This suggests that explicitly modeling temporal aesthetic dynamics is important for video aesthetic assessment. \ding{183} \textit{Failure cases and possible reasons.} We also observe several failure cases where \textit{Peak-End-Net} loses its advantage or produces less accurate predictions. A possible reason is that some videos contain subtle aesthetic cues that are difficult to infer from sparsely sampled frames alone, such as fine-grained semantic intent, narrative coherence, or subjective preference. In addition, when the visual quality is relatively uniform, the benefit of our temporal modeling may become less evident.

\bibliographystyle{ACM-Reference-Format}
\bibliography{sample-base}